\documentclass[10pt,journal,compsoc]{IEEEtran}
\pdfoutput=1
\usepackage{hyperref}
\hypersetup{
  pdfinfo={
    Title={title},
    Author={author},
  }
}
\usepackage{pdfpages}

\ifCLASSOPTIONcompsoc
  \usepackage[nocompress]{cite}
\else
  \usepackage{cite}
\fi

\ifCLASSOPTIONcompsoc
 \usepackage[caption=false,font=footnotesize,labelfont=sf,textfont=sf]{subfig}
\else
 \usepackage[caption=false,font=footnotesize]{subfig}
\fi

\usepackage{amsmath}
\usepackage{algorithmic}
\usepackage{array}
\usepackage{fixltx2e}
\usepackage{stfloats}
\usepackage{url}

\usepackage{times}
\usepackage{latexsym}
\usepackage{graphicx}
\usepackage{booktabs,multirow}
\usepackage{amssymb}
\usepackage{tabularx}
\usepackage{bm}
\usepackage{amsmath,mathtools}
\usepackage{amsfonts}
\usepackage{longtable}
\usepackage{wrapfig}
\usepackage[justification=centering]{caption}
\begin{document}
\bstctlcite{IEEEexample:BSTcontrol}
\title{Corruption Is Not All Bad: Incorporating Discourse Structure into Pre-training \\ via Corruption for Essay Scoring}

\ifCLASSOPTIONpeerreview
\begin{center} \bfseries EDICS Category: 3-BBND \end{center}
\fi

\author{Farjana Sultana Mim,~\IEEEmembership{}
        Naoya Inoue,~\IEEEmembership{}
        Paul Reisert,~\IEEEmembership{}
        Hiroki Ouchi and
        Kentaro Inui% <-this % stops a space
\IEEEcompsocitemizethanks{\IEEEcompsocthanksitem F. S. Mim is with the Graduate School of Information Sciences, Tohoku University, Japan. E-mail: farjana.mim59@gmail.com.
\IEEEcompsocthanksitem N. Inoue is with the Graduate School of Information Sciences, Tohoku University, Japan, and Riken Center for Advanced Intelligence Project, Japan. E-mail: naoya.inoue.lab@gmail.com. (currently in the Department of Computer Science, Stony Brook University, United States).
\IEEEcompsocthanksitem P. Reisert is with Riken Center for Advanced Intelligence Project, Japan, and the Graduate School of Information Sciences, Tohoku University, Japan. Email: paul.reisert@riken.jp
\IEEEcompsocthanksitem H. Ouchi is with Riken Center for Advanced Intelligence Project, Japan. Email: hiroki.ouchi@riken.jp.
\IEEEcompsocthanksitem K. Inui is with the Graduate School of Information Sciences, Tohoku University, Japan, and Riken Center for Advanced Intelligence Project, Japan. Email:inui@tohoku.ac.jp  }% <-this % stops an unwanted space
\thanks{}}

%%%%%%%%%%%%%%%%%%%%%%%%%%%%%%%%%%%%%%%%%%%%%%%%%%%%%%%%%%%%%%%%%%%%%%%%%%%%%%%%%%%%%%%%%
%%%%%%%%%%%%%%%%%%%%%%%%%%%%%%%%%%%%%%%%%%%%%%%%%%%%%%%%%%%%%%%%%%%%%%%%%%%%%%%%%%%%%%%%%
%%%%%%%%%%%%%%%%%%%%%%%%%%%%%%%%%%%%%%%%%%%%%%%%%%%%%%%%%%%%%%%%%%%%%%%%%%%%%%%%%%%%%%%%%

\IEEEtitleabstractindextext{%
\begin{abstract}
Existing approaches for automated essay scoring and document representation learning typically rely on discourse parsers to incorporate discourse structure into text representation. However, the performance of parsers is not always adequate, especially when they are used on noisy texts, such as student essays. 
In this paper, we propose an unsupervised pre-training approach to capture discourse structure of essays in terms of coherence and cohesion that does not require any discourse parser or annotation.
We introduce several types of token, sentence and paragraph-level corruption techniques for our proposed pre-training approach and augment masked language modeling pre-training with our pre-training method to leverage both contextualized and discourse information. 
Our proposed unsupervised approach achieves new state-of-the-art result on essay Organization scoring task.
\end{abstract}

}

\maketitle

\IEEEdisplaynontitleabstractindextext

\ifCLASSOPTIONpeerreview
\begin{center} \bfseries EDICS Category: 3-BBND \end{center}
\fi

% \fi
\IEEEpeerreviewmaketitle

%%%%%%%%%%%%%%%%%%%%%%%%%%%%%%%%%%%%%%%%%%%%%%%%%%%%%%%%%%%%%%%%%%%%%%%%%%%%%%%%%%%%%%%%%
%%%%%%%%%%%%%%%%%%%%%%%%%%%%%%%%%%%%%%%%%%%%%%%%%%%%%%%%%%%%%%%%%%%%%%%%%%%%%%%%%%%%%%%%%
%%%%%%%%%%%%%%%%%%%%%%%%%%%%%%%%%%%%%%%%%%%%%%%%%%%%%%%%%%%%%%%%%%%%%%%%%%%%%%%%%%%%%%%%%

\IEEEraisesectionheading{\section{Introduction}\label{sec:introduction}}

% \IEEEPARstart{D}{ocument} representation learning is essential for many natural language processing (NLP) tasks such as document classification (e.g., essay scoring and sentiment classification)~\cite{le2014distributed,liu2017learning,wu2018word,tang2015pte,mim2019unsupervised} and summarization.
% While learning approaches can be supervised, semi-supervised and unsupervised, recent studies have largely focused on unsupervised approaches in order to use large amounts of unlabeled texts and avoid time-consuming annotation procedures. 

\IEEEPARstart{A}{utomated} Essay Scoring (AES), the task of both grading and evaluating written essays using machine learning techniques, is an important educational application of natural language processing (NLP). 
% The aim of AES is to reduce the burden of manual evaluation for humans.
% in order to reduce the burden of manual evaluation for humans, is an important educational application of natural language processing (NLP). 
Since manual grading of student essays is extremely time consuming and requires lots of human efforts, AES systems are widely adopted for many large-scale writing assessments such as Graduate Record Examination (GRE)~\cite{attali2006automated}. Recent research in AES not only focuses on scoring overall quality (i.e., holistic scoring) of essays but also scoring a particular dimension of essay quality (e.g., Organization, Argument Strength, Style), in order to provide constructive feedback to learners~\cite{persing2010modeling, persing2013modeling, persing2014modeling, persing2015modeling, persing2016modeling, wachsmuth2016using, mathias2018thank, mim2019unsupervised}.

% The ultimate goal of AES is building a scoring system which not only assigns scores to essays but also provides constructive feedbacks to learners so that they can improve their writing skills.
% A major shortcoming of many AES systems is that they use holistic score of essays which is an overall score of essay quality. 
% Therefore, it is not clear how quality of different dimensions of an essay (e.g., Organization, Argument Strength, Style etc.) contributes to the score or if the score refers to a particular dimension. 
% As a consequence, holistic scoring limits the scope of providing elaborated and constructive feedbacks to learners.
% In order to address this issue, researchers have started scoring essays along a particular dimension~\cite{persing2010modeling, persing2013modeling, persing2014modeling, persing2015modeling, persing2016modeling, wachsmuth2016using, mathias2018thank, mim2019unsupervised}.

%~\cite{nadeem2019automated, liu2019automated, jin2018tdnn, cozma2018automated, farag2018neural ,taghipour2016neural}
% AES systems need to consider different aspects of an essay such as grammar, semantics, discourse, which makes the task more challenging. \cite{song2017discourse, somasundaran2014lexical}.

In general, an essay is a discourse where sentences and paragraphs are logically connected to each other to provide comprehensive meaning.
Conventionally, two types of connections have been discussed in the literature: \emph{coherence} and \emph{cohesion} \cite{halliday1994introduction}.
Coherence refers to the semantic relatedness among sentences and logical order of concepts and meanings in a text.
For example, \emph{``I saw Jill on the street. She was going home."} is coherent whereas \emph{``I saw Jill on the street. She has two sisters."} is incoherent.
Two types of coherence are well known in the literature: \emph{local coherence} and \emph{global coherence}.
Local coherence generally refers to how well-connected adjacent sentences are \cite{barzilay2008modeling} whereas global coherence represents the discourse relation among remote sentences to present the main idea of the text \cite{unger2006genre, zhang2011sentence}.
Cohesion refers to how well sentences and paragraphs in a text are linked by means of linguistic devices.
Examples of these linguistic devices include conjunctions such as discourse indicators (DIs) (e.g., \emph{``because" and ``for example"}), coreference (\emph{e.g., ``he" and ``they"}), substitution, ellipsis, etc.

% In document-level NLP tasks that need a precise assessment of documents, it is crucial to encode such discourse structure into a document representation.
% One example of these tasks is essay scoring, especially when scoring essays along a particular dimension.
% One such dimension is \emph{Organization}, which refers to how good an essay structure is. % where ``well-organized" essays logically develop arguments and state positions by supporting them~\cite{persing2010modeling}.

For the precise assessment of overall essay quality or some dimensions of an essay, it is crucial to encode such discourse structure (i.e., coherence and cohesion) into an essay representation.
One such dimension is \emph{Organization}, which refers to how good an essay structure is.
Essays with a high Organization score have the structure where writers introduce a topic first, state their position regarding the topic, support their position by providing reasons and then conclude often by stating their position again~\cite{persing2010modeling}.

% In our previous study~\cite{mim2019unsupervised}, we proposed an unsupervised method to capture discourse structure in terms of coherence and cohesion for document representation. 
% % The idea was to first pre-train a document encoder with unlabeled data which would learn to distinguish between original and corrupted documents. 
% % Then we used the pre-trained encoder to obtain feature vectors of essays for Organization and Argument Strength score prediction. After that, the feature vectors were mapped to scores by regression. 
% Our experimental results showed that the proposed pre-training strategy is able to improve the performance of essay Organization and Argument Strength scoring.

% However, the characteristics of these two individual scoring are quite different. 
% Essay with high Organization score helps readers understand writer's position as well as supporting claims properly whereas in an essay with high score on Argument Strength, the writer's arguments to support his/her position would convince most of the readers~\cite{persing2010modeling, persing2015modeling}. Therefore, in this study we have decided to focus on one particular dimension: \emph{Organization}. 

\begin{figure*}[ht!]
\centering
  \includegraphics[height=75mm, width=17cm]{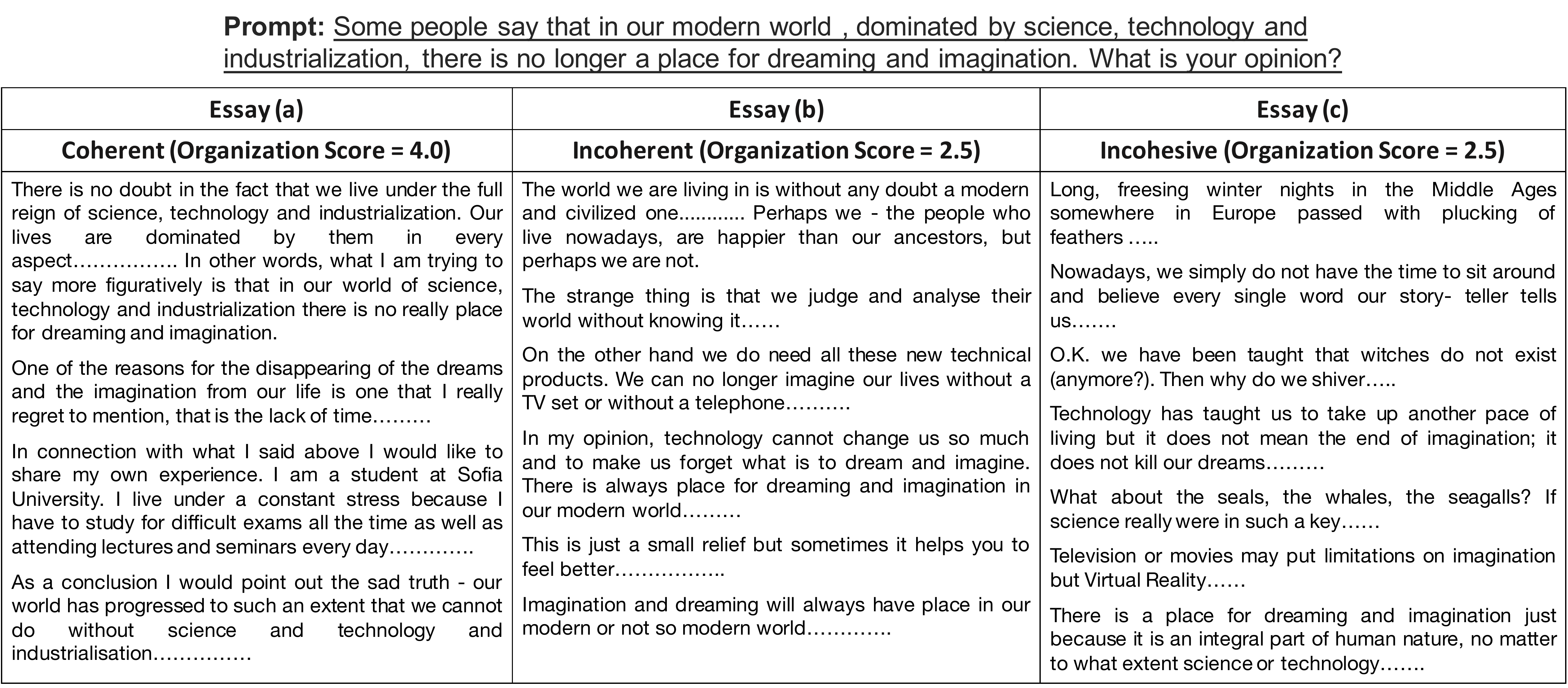}
  \captionsetup{justification=centering}
  \caption{Example of coherent/cohesive and incoherent/incohesive essays with their Organization score.}
  \label{Fig:essay_example}
\end{figure*}

An example of the relation between coherence, cohesion and an essay's Organization is shown in Figure~\ref{Fig:essay_example}.
The high-scored essay (i.e., Essay (a) with Organization score 4) first states its position regarding the prompt and then provides several reasons to strengthen the claim.
It is considered coherent because it follows a logical order that makes the writer's position and arguments very clear. 
However, Essay (b) is not clear on its position and what it is arguing about.
The third paragraph gives a vibe that the writer is supporting the prompt but then the fourth paragraph provides a clear statement that the writer is opposing the prompt.
Therefore, it can be considered incoherent since it lacks logical sequencing.

Furthermore, Essay (a) has cohesive markers (e.g., ``in connection with", ``as a conclusion") at the beginning of the paragraphs which helps the reader understand the flow of ideas throughout the essay. 
Therefore, it is considered as a cohesive essay.
However, Essay (c) should have some cohesive markers at the beginning of fifth paragraph (e.g., ``moreover", ``besides") and sixth paragraph (e.g., ``therefore", ``hence") to connect the ideas between paragraphs. Besides, there is no cohesive marker at the beginning of the last paragraph (e.g., ``in conclusion") to indicate that the author is summing up his/her opinions which makes the last paragraph somewhat disconnected from former paragraphs. Due to the absence of these cohesive markers, it is difficult to understand the arguments and connections between them.
Therefore, the essay is considered as an incohesive essay.

% Previous studies of unsupervised document representation learning mainly focused on capturing word similarity, word and sentence dependencies, semantic information and hierarchical structure of documents~\cite{le2014distributed,liu2017learning,wu2018word,amiri2019vector, Ionescu2019VectorOL, gupta2020psif, chang2019language}. 

Although discourse is one of the most important aspects of documents, less attention has been paid to capturing discourse structure in an unsupervised manner for document representation. Most of the works that encapsulate discourse structure into document representation are dependent on Rhetorical Structure Theory (RST) based or argumentative parser and annotations~\cite{stab2014annotating, stab2014identifying, mann1988rhetorical}.
However, such annotations are costly and during parsing, the parsers generally consider that the text is well-written which is not always true, specially in case of student essays that comprise different types of flaws (e.g., grammatical, spelling, discourse etc.).
To sum up, using parsers for document representation has its own limitations~\cite{ji2017neural}, specially when used on poorly written text and it has not yet been explored how long-range discourse dependencies can be included in text embedding in an unsupervised way without any parser or annotation.

Recent advances in language model (LM) pre-training has inspired researchers to use contextualized language representations for different document-level downstream tasks of NLP, including essay scoring. Several document-level tasks such as document classification, summarization~\cite{adhikari2019docbert, zhang2019hibert, xu2020discourse} as well as essay scoring~\cite{steimeltowards, liu2019automated, nadeem2019automated} achieved state-of-the-art performance by leveraging pre-trained language models. 
It should be mentioned that many of these tasks obtained only the sentence or text block representation from the pre-trained language models instead of the whole document representation and later joined them using some complex architecture, because Transformer-based~\cite{vaswani2017attention} pre-trained models (e.g., BERT, RoBERTa~\cite{devlin2018bert, liu2019roberta}) is infeasible to process long document due to the token constraints (they accept up to 512 tokens).
Furthermore, due to the self-attention operation of Transformer, processing long documents is very expensive.
The recent work of Beltagy et al. \cite{Beltagy2020Longformer} addressed these limitations and introduced Transformer-based model \emph{Longformer} which is suitable for processing long documents. % Their pre-trained Longformer model outperformed renowned RoBERTa on various long document tasks.
However, long-range discourse dependencies are not well captured by the pre-tranied language models \cite{xu2020discourse} because of the token and sentence level pre-training (not document level).
% and less attention has been paid to capturing such discourse structure in an unsupervised manner for text representation (see Section~\ref{sec:related_work} and \ref{sec:model} for more details).

% These pre-trained language models achieved state-of-the-art result on different document-level downstream tasks of NLP, including essay scoring~\cite{Beltagy2020Longformer, adhikari2019docbert, zhang2019hibert, xu2020discourse, steimeltowards, liu2019automated}.

In this paper, we propose an unsupervised method that enhances a document encoder to capture discourse structure of essay Organization in terms of cohesion and coherence (Section~\ref{sec:model},\ref{sec:proposed_method}).
We name our unsupervised technique as \emph{Discourse Corruption (DC)} pre-training.
We introduce several types of token, sentence and paragraph level corruption strategy to artificially produce ``badly-organized" (incoherent/incohesive) essays.
We then pre-train a document encoder which learns to discriminate between original (coherent/cohesive) and corrupted (incoherent/incohesive) essays. 

We augment Longformer~\cite{Beltagy2020Longformer}, a strong document encoder pre-trained with a Masked Language Modeling (MLM) objective, with the DC pre-training in order to utilize both contextual and discourse information of essays.
We expect that the MLM objective will capture the transition of ideas at local level (e.g., word or sentence level) while our DC pre-training will capture the transition of ideas at global level (e.g., paragraph), and the combination of these two strategies will successfully capture the overall Organization structure of an essay.
To the best of our knowledge, we are the first to attach discourse-aware pre-training on top of MLM pre-training.
The advantage of our approach is that it is unsupervised and does not require any parser or annotation.
Our proposed strategy outperforms a baseline model by a significant margin, and we achieve new state-of-the-art result for essay Organization scoring (Section~\ref{sec:setup},\ref{Sec:results}).

\section{Related Work}
\label{sec:related_work}
The focus of this study is the unsupervised encapsulation of discourse structure into document representation for essay Organization scoring.
In this section, we briefly review the previous works on automated essay scoring, unsupervised document representation learning and document representation learning using pre-trained language models.

% A popular approach for document representation is the use of fixed-length features such as bag-of-words (BOW) and bag-of-ngrams due to their simplicity and highly competitive results~\cite{wang2012baselines}.
% However, such approaches fail to capture the semantic similarity of words and phrases since they treat each word or phrase as a discrete token.

\subsection{Automated Essay Scoring}
AES research generally follows two lines of approaches: feature-engineering approach and deep neural network (DNN) based approach. 
Traditional AES research utilizes handcrafted features in a supervised regression or classification setting to predict the score of essays~\cite{larkey1998automatic, attali2006automated, chen2013automated, phandi2015flexible, persing2010modeling, persing2015modeling, wachsmuth2016using}.
Recent studies of AES adopt DNN based approaches which have shown very promising results~\cite{taghipour2016neural, alikaniotis2016automatic, dong2016automatic, dong2017attention, riordan2017investigating, farag2018neural, zhang2018co, wang2018automatic, cummins2018neural}.

% A major shortcoming of many AES systems is that they use holistic score of essays which is an overall score of essay quality. 
% Therefore, it is not clear how quality of different dimensions of an essay (e.g., Organization, Argument Strength, Style etc.) contributes to the score or if the score refers to a particular dimension. 
% As a consequence, holistic scoring limits the scope of providing elaborated and constructive feedbacks to learners.
% In order to address this issue, researchers have started scoring essays along a particular dimension

% Although majority of the AES systems use holistic scoring scheme

A major shortcoming of many of the AES systems is that they use holistic score of essays~\cite{phandi2015flexible, alikaniotis2016automatic, taghipour2016neural, dong2017attention, wang2018automatic}.
Holistic scoring schemes limit the scope of providing constructive feedback to learners since from the score it is not clear how different dimensions of essay quality (e.g., Organization, content) are summarized into a single score or if the score refers to a particular dimension.
In order to address this issue, researchers have focused on scoring specific dimensions of essay such as organization, argument strength~\cite{persing2010modeling, persing2015modeling, wachsmuth2016using}, thesis clarity~\cite{persing2013modeling}, relevance to prompt~\cite{higgins2004evaluating, persing2014modeling}, stance~\cite{persing2016modeling}, style~\cite{mathias2018thank} etc. Discourse coherence, one of the important dimensions of essay quality, has also been exploited for essay assessment.
Mesgar et al.~\cite{mesgar2018neural} used an end-to-end local coherence model for the assessment of essays that encodes semantic relations of two adjacent sentences and their pattern of changes throughout the text. %where the essays are annotated with holistic scores.
Farag et al. \cite{farag2018neural} evaluated the robustness of neural AES model and showed that neural AES model is not well-suited for capturing adversarial input of grammatically correct but incoherent sequences of sentences.
Therefore, they developed a neural local coherence model and jointly trained it with a state-of-the-art AES model to build an adversarially robust AES system.
However, these works utilized the particular essay quality ``coherence" for the assessment of overall essay quality (holistic scoring). In this work, we capture discourse cohesion and coherence in an unsupervised way for scoring a specific dimension of essay i.e., Organization.

Recently, pre-trained deep language representation models have fascinated the NLP community by achieving state-of-the-art result on various downstream tasks of NLP, including essay scoring. 
One of the widely used masked language models is \emph{BERT: Bidirectional Encoder Representations from Transformers} \cite{devlin2018bert} which was trained with MLM objective i.e., predicting the masked tokens in the text.
%Pre-trained language models have influenced the essay scoring area as well. 
Several essay scoring tasks achieved state-of-the-art performance by leveraging BERT \cite{devlin2018bert}. Steimel et al. \cite{steimeltowards} fine tuned BERT and achieved state-of-the-art result for content scoring of essays. 
Liu et al. \cite{liu2019automated} proposed a two stage learning framework (TSLF) that integrates both end-to-end neural AES model as well as feature-engineered model and achieved state-of-the-art performance on holistic scoring of essays. 
In their framework, sentence embeddings are obtained using the pre-trained BERT model.
They also incorporated Grammar Error Correction (GEC) system into their AES model and added adversarial samples to the original dataset which led to performance gain. 
Nadeem et al. \cite{nadeem2019automated} used existing discourse-aware models and tasks from literature to pre-train AES models for holistic scoring of essay. They used natural language inference and discourse marker prediction tasks as their pre-training objectives as well as contextualized BERT embeddings, hypothesizing that the next sentence prediction task of BERT would capture discourse coherence. Their results also showed that contextualized embeddings from BERT performs better than other two pre-training tasks.

However, all these studies consider holistic scores where it is unclear which criteria of the essay the score considers. We are the first to show how Transformer-based~\cite{vaswani2017attention} architecture with MLM pre-training performs on the assessment of a specific dimension of essay i.e. essay Organization scoring.
Persing et al. \cite{persing2010modeling} annotated essays with Organization scores and established a baseline model for this scoring. They employed heuristic rules utilizing various DIs, words, and phrases to capture the discourse function labels of sentences and paragraphs of an essay. Then those function labels were exploited by various techniques such as sequence alignment, alignment kernels, string kernels, for the prediction of Organization score.
Later, Wachsmuth et al. \cite{wachsmuth2016using} achieved state-of-the-art performance on Organization scoring by utilizing argumentative features such as sequence of argumentative discourse units (ADU) (e.g., \emph{(conclusion, premise, conclusion)}, \emph{(None, Thesis})), frequencies of ADU types, etc.
In addition to the argumentative features, they also used sequences of paragraph discourse functions of Persing et al. \cite{persing2010modeling} as well as sentiment flows, relation flows, POS n-grams, frequency of tokens in training essays etc.
Then a simple supervised regression model is applied for scoring.
However, their work use an argument parser to obtain ADUs and we would like to overcome that parser bottleneck.

It should be noted that our proposed unsupervised DC pre-training was first introduced in our previous works \cite{mim2019unsupervised, mim2019unsupervisedanlp}.
The document representation obtained from DC pre-training was used for essay Organization and Argument Strength scoring.
However, in this study, we only focus on essay Organization scoring.
In this work, we present several new corruption techniques in addition to our previous corruption strategies~\cite{mim2019unsupervised} to capture the Organization structure of essays.
Besides, in contrast to our previous research, in this study we use a Transformer-based model pre-trained with MLM objective as our document encoder and augment our DC pre-training on top of it.
To elaborate, in this paper, we extend our previous research by introducing new corruption techniques and by enhancing a document encoder with our DC pre-training to capture discourse structure of essay Organization.

\subsection{Unsupervised Document Representation Learning}
Several unsupervised methods for document representation learning have been introduced in recent years~\cite{le2014distributed, wu2018word, Ionescu2019VectorOL, gupta2020psif}.
However, less studies have been conducted on unsupervised learning of discourse-aware text representation.
% Although these unsupervised techniques have been proven useful for several document classification and regression tasks, their focus is not on capturing the discourse structure of documents.
% However, less study has been conducted on capturing discourse structure of documents.
One of the studies that illustrated the role of discourse structure for document representation is the study by Ji and Smith \cite{ji2017neural} who implemented a discourse structure (defined by RST) \cite{mann1988rhetorical} aware model and showed that their model improves text categorization performance (e.g., sentiment classification of movies and Yelp reviews, and prediction of news article frames).
The authors utilized an RST-parser to obtain the discourse dependency tree of a document and then built a recursive neural network on top of it. 
The issue with their approach is that texts need to be parsed by an RST parser and the parsing performance of RST is not always adequate, specially when used on noisy text.
Furthermore, the performance of RST parsing is dependent on the genre of documents~\cite{ji2017neural}.

\subsection{Pre-trained Language models and Document Representation Learning}
% Recently, pre-trained deep language representation models have fascinated the NLP community by achieving state-of-the-art result on various downstream tasks of NLP. 
% One of the popular masked language models used by different document-level tasks is BERT: Bidirectional Encoder Representations from Transformers \cite{devlin2018bert} where the training objective is to predict the masked tokens in the text.
Lately, Tansforemer-based pre-trained models have achieved significant performance gain in different document-level downstream tasks of NLP.
Adhikari et al. \cite{adhikari2019docbert} first instigated the use of pre-trained deep contextualized models for document representation learning. They fine-tuned BERT~\cite{devlin2018bert} for several document classification tasks and demonstrated that knowledge can be distilled from BERT to small bidirectional LSTMs which provides competitive results at a low computational expense.

Chang et al. \cite{chang2019language} proposed methods for pre-training hierarchical document representations that generalize and extend the pre-training method of ELMo \cite{peters2018deep} and BERT \cite{devlin2018bert} respectively. In their approach, LSTM-based architecture consider a document as sequences of text blocks, each block comprising a sequence of tokens, where the text blocks are basically sentences or paragraphs.
Zhang et al. \cite{zhang2019hibert} presented a strategy to pre-train hierarchical bidirectional transformer encoders for document representation. They randomly masked sentences of documents and predicted those masked sentences with their proposed architecture, a hierarchical fusion of  Transformer-based \cite{vaswani2017attention} sentence and document encoder.

A recent work by Beltagy et al. \cite{Beltagy2020Longformer} pointed out the attention mechanism and token constraints of Transformer-based \cite{vaswani2017attention} masked language models for long document representation. To mitigate these problems, they introduced \emph{Longformer: The Long-Document Transformer} model with an attention mechanism that scales linearly with the sequence length, hence being suitable for processing long documents. They pre-trained the Longformer with MLM objective, continuing from the RoBERTa \cite{liu2019roberta}  released checkpoint and added extra position embeddings to support long sequence of tokens. 
The pre-trained Longformer outperformed renowned RoBERTa on various long document tasks.

One recent study by Xu et al. \cite{xu2020discourse} utilized pre-trained language model for capturing discourse structure of documents. They constructed a discourse-aware
neural extractive summarization model \emph{DISCOBERT}. DISCOBERT encodes RST-based discourse unit (a sub-sentence phrase) instead of sentence using BERT. Then Graph Convolutional Network is used to create discourse graphs based on RST trees and coreference mentions. However, this work is dependent on the RST discourse parser and as we mentioned earlier,  we would like to overcome that parser bottleneck.

\begin{figure*}[!ht]
  \centering
  \includegraphics[width=13.5cm, height=6cm]{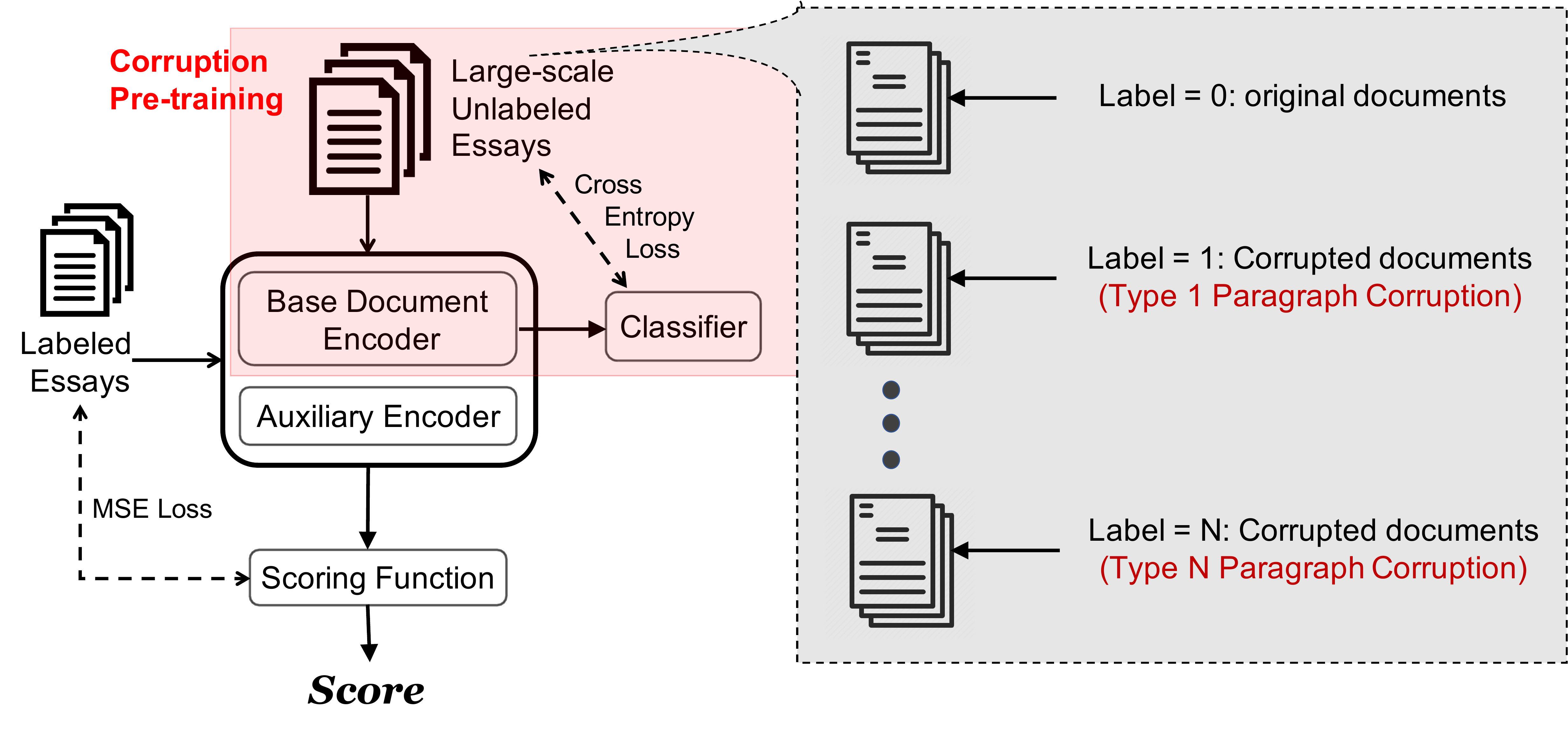}
  \captionsetup{margin = 35pt}
  \caption{Proposed DC pre-training for unsupervised learning of discourse-aware text representation utilizing original and artificially corrupted documents and the use of the discourse-aware pre-trained model for essay scoring.}
  \label{Fig:idea}
\end{figure*}

\begin{figure*}[!ht]
  \centering
  \includegraphics[width=17cm, height=6.4cm]{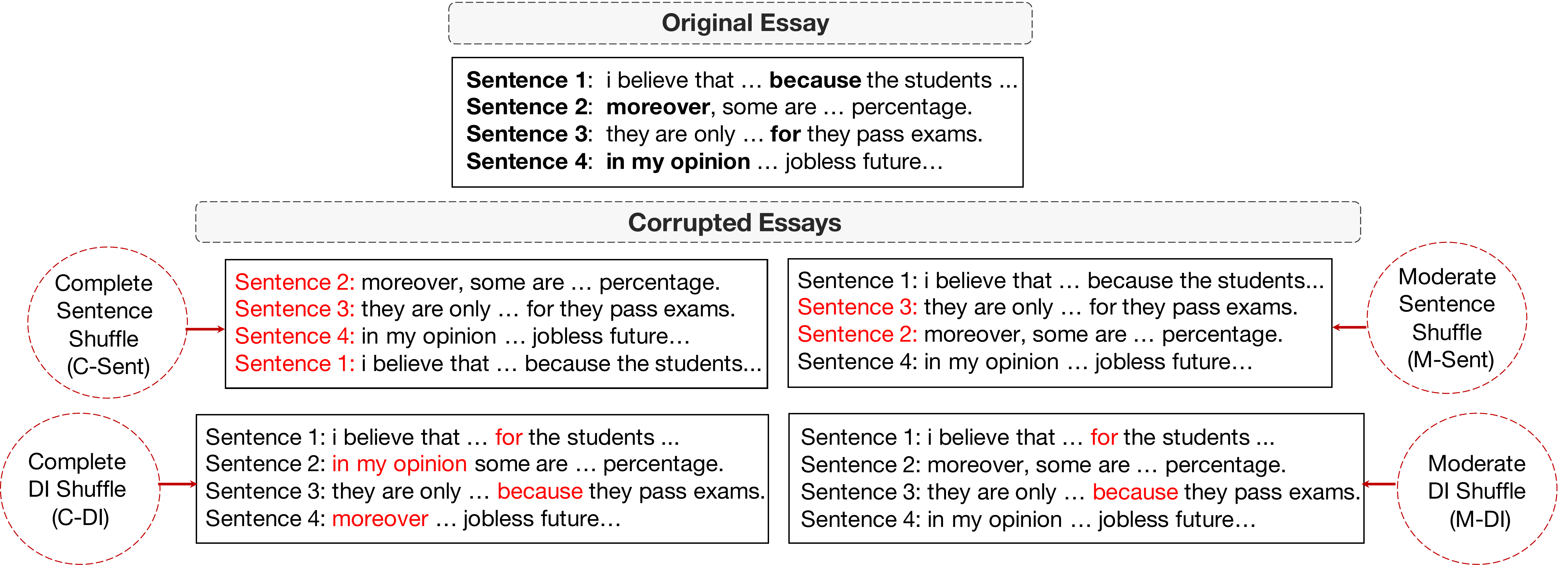}
  \caption{Example of different types of Sentence and Discourse Indicator Corruption}
  \label{Fig:sc_dic_example}
\end{figure*}

%%%%%%%%%%%%%%%%%%%%%%%%%%%%%%%%%%%%%%%%%%%%%%%%%%%%%%%%%%%%%%%%%%%%%%%%%%%%%%%%%%%%%%%%%
%%%%%%%%%%%%%%%%%%%%%%%%%%%%%%%%%%%%%%%%%%%%%%%%%%%%%%%%%%%%%%%%%%%%%%%%%%%%%%%%%%%%%%%%%
%%%%%%%%%%%%%%%%%%%%%%%%%%%%%%%%%%%%%%%%%%%%%%%%%%%%%%%%%%%%%%%%%%%%%%%%%%%%%%%%%%%%%%%%%

\section{Model Architecture}
\label{sec:model}

\subsection{Overview}
\label{sec:overview}
Our model consists of (i) a base document encoder, (ii) an auxiliary encoder, and (iii) a scoring function.
The base document encoder produces a vector representation ${\bf h}^\text{base}$ by capturing a sequence of words in each essay.
The auxiliary encoder captures additional essay-related information and produces a vector representation ${\bf h}^\text{aux}$. 

% Specifically, these encoders first produce the representations, ${\bf h}^\text{base}$ and ${\bf h}^\text{aux}$.
Then, these representations are concatenated into one vector, which is mapped to a feature vector ${\bf z}$.
\begin{equation}
\label{eq:overview}
    {\bf z} = {\rm tanh}({\bf W} \cdot [{\bf h}^\text{base}; {\bf h}^\text{aux}]) \:\:,
\end{equation}
where ${\bf W}$ is a weight matrix.
Finally, we use the following scoring function to map ${\bf z}$ to a scalar value by the sigmoid function.
\begin{equation} \nonumber
    y = {\rm sigmoid}({\bf w} \cdot {\bf z} + b) \:\:,\nonumber
\end{equation}
where {\bf w} is a weight vector, $b$ is a bias value, and $y$ is a score in the range of $[0, 1]$.
In the following subsections, we describe the details of each encoder.

\subsection{Base Document Encoder}
\label{sec:doc_enc}
The base document encoder produces a document representation ${\bf h}^\text{base}$ in Equation~\ref{eq:overview}.
For the base document encoder, we use the pre-trained Longformer model \cite{Beltagy2020Longformer}.
% Longformer is pre-trained with masked language modeling (MLM) where the objective is to predict randomly masked tokens.

Longformer is a Transformer-based \cite{vaswani2017attention} model with modified attention mechanism. 
Longformer's attention mechanism scales linearly with the input sequence length, making it easy for processing long documents.
The attention mechanism of Longformer combines a sliding windowed self-attention for capturing local-context and a task specific global attention. In this attention operation, if the sliding window size is $w$, then each token will attend to $\frac{1}{2}w$ token on each side and a token with a global attention will attend to all the tokens across the sequence and all the tokens in the sequence will attend to it as well.
Longformer is pre-trained with the MLM objective, continued from the RoBERTa released checkpoint.
During pre-training, Longformer's attention mechanism is used as a drop-in replacement for the self-attention mechanism of Transformer-based RoBERTa \cite{liu2019roberta}.
Specifically, RoBERTa's self-attention is replaced by Longformer's attention.
Longformer can process much longer documents by accepting up to 4096 tokens whereas other pre-trained models like BERT \cite{devlin2018bert} or RoBERTa \cite{liu2019roberta} only accepts up to 512 tokens.
Since the Transformer architecture \cite{vaswani2017attention} is well-known and widely used in NLP, we will omit the exhaustive review of it. Instead, we would present a brief overview of how Longformer is used in our essay scoring model.

Given an input essay of $N$ tokens $t_{1:N} = (t_1, t_2, \cdots, t_N)$, special tokens are inserted at the beginning and the end of the essay, finally the input essay of $N$ tokens being $t_{0:N+1} = (\text{[CLS]}, t_1, t_2, \cdots, t_N, \text{[EOS]})$.
Then, taking $t_{0:N+1}$ as input, the Longformer model produces a sequence of contextual representations ${\bf h}_{0:N+1} = ({\bf h}_0, {\bf h}_1, \cdots, {\bf h}_{N+1})$. Note that, we obtain the representation from the second-to-last layer of Longformer.
\begin{equation} \nonumber
{\bf h}_{0:N+1} = \text{Longformer}(t_{0:N+1}) \:\: ,
\end{equation}

After that, we use a mean-over-time layer where taking ${\bf h}_{0:N+1}$ as input, it produces a vector averaged over the sequence.
\begin{equation}
\label{eq:base_doc}
% {\bf h}^\text{mean} = \frac{1}{N} \sum^N_{n = 1} {\bf h}_n \:\:.
h^\mathrm{mean} = \frac{1}{N+2} \sum^{N+1}_{n=0} {\bf h}_n \:\:.
%{\bf h}^\text{mean} = \text{moe}({\bf h}_{1:T}) = \frac{1}{T} \sum_{t \in [1,T]} {\bf h}_t \:\:.
\end{equation}

\noindent
We use this resulting vector as the base document representation, i.e. ${\bf h}^\text{base} = {\bf h}^\text{mean}$.

% % This model uses three types of layers: an embedding layer, a Bi-directional Long Short-Term Memory (BiLSTM) \cite{lstm} layer and a mean-over-time layer.
% Given the input essay of $T$ words $w_{1:T} = (w_1, w_2, \cdots, w_T)$, the embedding layer (Emb) produces a sequence of word embeddings ${\bf w}_{1:T} = ({\bf w}_1, {\bf w}_2, \cdots, {\bf w}_T)$.
% \begin{equation}
% {\bf w}_{1:T} = \text{Emb}(w_{1:T}) \:\:, \nonumber
% \end{equation}

% \noindent
% where each word embedding is a $d^\text{word}$ dimensional vector, i.e. ${\bf w}_i \in \mathbb{R}^{d^\text{word}}$.

% Then, taking ${\bf x}_{1:T}$ as input, the BiLSTM layer produces a sequence of contextual representations ${\bf h}_{1:T} = ({\bf h}_1, {\bf h}_2, \cdots, {\bf h}_T)$.
% \begin{equation} \nonumber
% {\bf h}_{1:T} = \text{BiLSTM}({\bf x}_{1:T}) \:\: ,
% \end{equation}

% \noindent
% where each representation ${\bf h}_i$ is $\mathbb{R}^{d^\text{hidden}}$.

% Finally, taking ${\bf h}_{1:T}$ as input, the mean-over-time layer produces a vector averaged over the sequence.
% \begin{equation}
% \label{eq:base_doc}
% {\bf h}^\text{mean} = \frac{1}{T} \sum^T_{t = 1} {\bf h}_t \:\:.
% %{\bf h}^\text{mean} = \text{moe}({\bf h}_{1:T}) = \frac{1}{T} \sum_{t \in [1,T]} {\bf h}_t \:\:.
% \end{equation}

% \noindent
% We use this resulting vector as the base document representation, i.e. ${\bf h}^\text{base} = {\bf h}^\text{mean}$.

\subsection{Auxiliary Encoder}
\label{sec:ae}
The auxiliary encoder produces a representation of a sequence of paragraph function labels ${\bf h}^\text{aux}$ in Equation~\ref{eq:overview}.

% \subsubsection{Paragraph Function Encoder (PFE).}

Each paragraph in an essay plays a different role.
For instance, the first paragraph tends to introduce the topic of the essay, and the last paragraph tends to sum up the whole content and make some conclusions.
Here, we capture such paragraph functions.

Specifically, we obtain paragraph function labels of essays using Persing et al.'s \cite{persing2010modeling} heuristic rules.\footnote{See \url{http://www.hlt.utdallas.edu/~persingq/ICLE/orgDataset.html} for further details.}
Persing et al. \cite{persing2010modeling} specified four paragraph function labels: Introduction (\textbf{I}), Body (\textbf{B}), Rebuttal (\textbf{R}) and Conclusion (\textbf{C}).
We represent these labels as vectors and incorporate them into our model.
Our auxiliary encoder that encodes paragraph function labels consists of two modules, an embedding layer and a  Bi-directional Long Short-Term Memory (BiLSTM) \cite{lstm} layer.

We assume that an essay consists of $M$ paragraphs, and the $i$-th paragraph has already been assigned a function label $p_i$.
Given the sequence of paragraph function labels of an essay $p_{1:M} = (p_1, p_2, ..., p_M)$, the embedding layer (Emb$^\text{para}$) produces a sequence of label embeddings ${\bf p}_{1:M} = ({\bf p}_1, {\bf p}_2, \cdots, {\bf p}_M)$.
\begin{equation}\nonumber
    {\bm p}_{1:M} = \text{Emb}^\text{para}(p_{1:M}),
\end{equation}
where each embedding ${\bm p}_i$ is $\mathbb{R}^{d^\text{para}}$.
Note that each embedding is randomly initialized and learned during training.

Then, taking ${\bm p}_{1:M}$ as input, the BiLSTM layer produces a sequence of vector representations ${\bf h}_{1:M} = ({\bf h}_1, {\bf h}_2, \cdots, {\bf h}_M)$.
\begin{equation}\nonumber
    {\bf h}_{1:M} = \text{BiLSTM}({\bm p}_{1:M}),
\end{equation}
where ${\bf h}_i$ is $\mathbb{R}^{d^\text{aux}}$.

We use the last hidden state ${\bf h}_M$ as the paragraph function label sequence representation, i.e. ${\bf h}^\text{aux} = {\bf h}_M$.

%%%%%%%%%%%%%%%%%%%%%%%%%%%%%%%%%%%%%%%%%%%%%%%%%%%%%%%%%%%%%%%%%%%%%%%%%%%%%%%%%%%%%%%%%
%%%%%%%%%%%%%%%%%%%%%%%%%%%%%%%%%%%%%%%%%%%%%%%%%%%%%%%%%%%%%%%%%%%%%%%%%%%%%%%%%%%%%%%%%
%%%%%%%%%%%%%%%%%%%%%%%%%%%%%%%%%%%%%%%%%%%%%%%%%%%%%%%%%%%%%%%%%%%%%%%%%%%%%%%%%%%%%%%%%

\section{Proposed Pre-training Method}
\label{sec:proposed_method}
\subsection{Overview}
Figure~\ref{Fig:idea} summarizes our proposed DC pre-training method.
First, we pre-train the base document encoder (Section~\ref{sec:doc_enc}) to distinguish between original and their artificially corrupted documents.
This pre-training is motivated by the following hypotheses: (i) artificially corrupted incoherent/incohesive documents lack logical sequencing, (ii) moderately corrupted documents have better logical sequencing than highly corrupted documents and (iii) training a base document encoder to differentiate between original and their different types of artificially corrupted documents makes the encoder logical sequence-aware, in other words, discourse-aware.
Based on these hypotheses, we train a base document encoder on the original and their artificially corrupted documents.

The pre-training is done in two steps.
First, we pre-train the document encoder with large-scale unlabeled essays of different corpus.
Second, we fine-tune the encoder on the unlabeled essays of target corpus (essay Organization scoring corpus).
We expect that this fine-tuning alleviates the domain mismatch between the large-scale essays and target essays (e.g., essay length).
Finally, the pre-trained encoder is then re-trained on the annotations of essay scoring task in a supervised manner.

Note that, our base document encoder (i.e., Longformer) is already pre-trained with the MLM objective, where the aim is to predict randomly masked tokens in a sequence.
%Therefore, we basically add the Corruption Pre-training on top of MLM pre-training. 
We expect that MLM pre-training would capture local-context while our DC pre-training will capture the long-range dependencies effective for essay Organization scoring.
% From the two sizes of pre-trained Longformer model, we use Longformer-base model.

\subsection{Corruption Strategies}
We would like to produce ``badly organized'' essays with our corruption techniques so that the encoder can learn the difference between good and bad discourse. Note that, essays are not only scored as high or low but throughout a range of score which means that, there is Organization structure which is moderately good/bad.
Therefore, in addition to the high corruption techniques, we introduce several types of moderate corruption techniques in order to produce ``moderately bad'' Organization of essays.

We categorize our corruption strategies into 3 groups: (1) \emph{sentence}, (2) \emph{discourse indicator (DI)} and (3) \emph{paragraph} corruption. 
Each group has several types of corruption scheme. 
We discuss the details of each corruption strategy in the following subsections.

\subsubsection{Sentence Corruption (SC)}

\newcommand{\csslong}{Complete Sentence Shuffle}
\newcommand{\msslong}{Moderate Sentence Shuffle}
\newcommand{\css}{C-Sent}
\newcommand{\mss}{M-Sent}

This group has 2 different types of corruption.
In \emph{\csslong{}} (\css{}), all the sentences of a document is shuffled. 
In \emph{\msslong{}} (\mss{}), only subset of the sentences of a document are shuffled.
Specifically, we randomly select two sentences from a document and shuffle all the sentences between them, including those two sentences as well.
Figure~\ref{Fig:sc_dic_example} shows an example of \css{} and \mss{}.

\begin{figure*}[!ht]
  \centering
  \includegraphics[width=17cm, height=6.5cm]{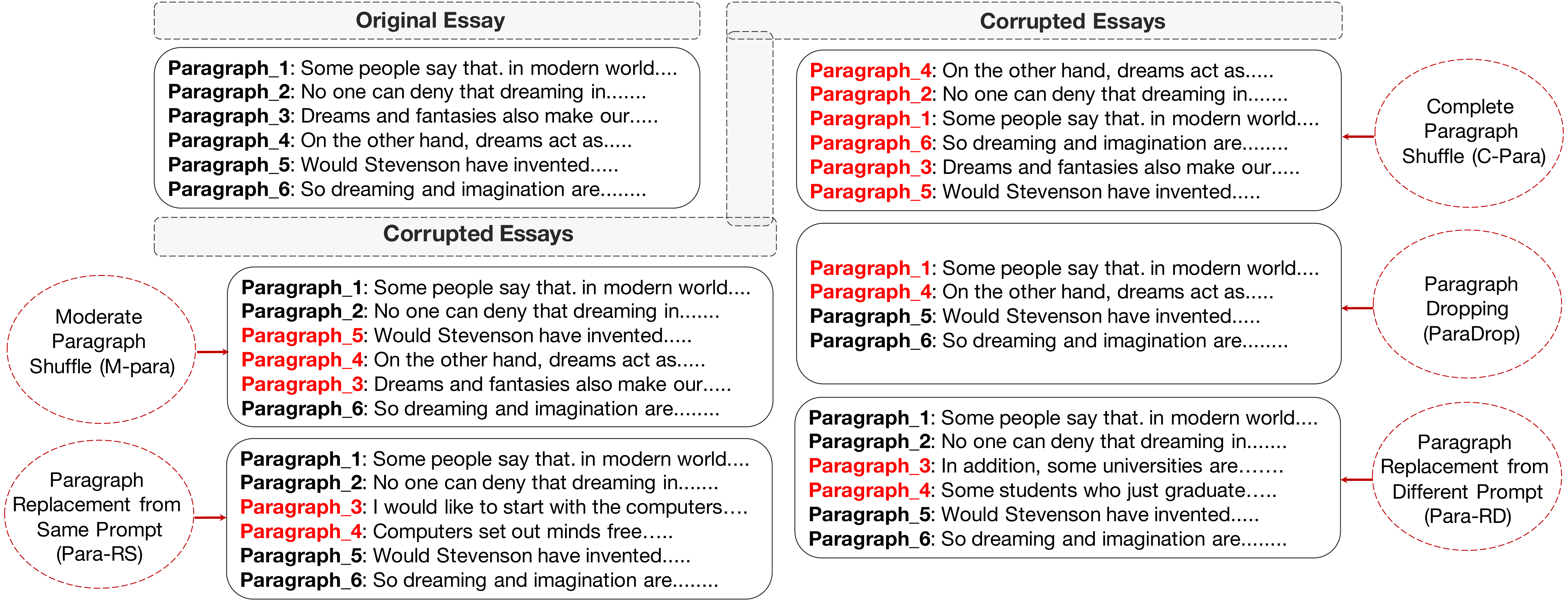}
  \caption{Example of different types of Paragraph Corruption}
  \label{Fig:pc_example}
\end{figure*}

\newcommand{\cdislong}{Complete Discourse Indicator Shuffle}
\newcommand{\mdislong}{Moderate Discourse Indicator Shuffle}
\newcommand{\cdis}{C-DI}
\newcommand{\mdis}{M-DI}

\subsubsection{Discourse Indicator Corruption (DIC)}
We corrupt DIs since they represent logical connection between sentences.
For example, \emph{``Mary did well although she was ill"} is logically connected, but \emph{``Mary did well but she was ill."} and \emph{``Mary did well. She was ill."} lack logical sequencing because of improper and lack of DI usage, respectively.

We perform two types of DI corruption.
In \emph{\cdislong{}} (\cdis{}), we shuffle all the discourse indicators of a document.
In \emph{\mdislong{}} (\mdis{}), we shuffle randomly selected 50\% of all the unique DIs of a document.
Figure~\ref{Fig:sc_dic_example} shows an example of \cdis{} and \mdis{}.

\newcommand{\cpslong}{Complete Paragraph Shuffle}
\newcommand{\mpslong}{Moderate Paragraph Shuffle}
\newcommand{\pdlong}{Paragraph Drop}
\newcommand{\prsplong}{Paragraph Replacement from Same Prompt}
\newcommand{\prdplong}{Paragraph Replacement from Different Prompt}
\newcommand{\cps}{C-Para}
\newcommand{\mps}{M-Para}
\newcommand{\pd}{ParaDrop}
\newcommand{\prsp}{Para-RS}
\newcommand{\prdp}{Para-RD}

\subsubsection{Paragraph Corruption (PC)}
How ideas are transmitted throughout the paragraphs of an essay determines how good its Organization structure is. For example, coherent essays have paragraph sequences like \emph{Introduction-Body-Conclusion} to provide a logically consistent meaning of the text. 
Therefore, we conduct five types of paragraph corruption, as illustrated in Figure~\ref{Fig:pc_example}.

In \emph{\cpslong{}} (\cps{}), we randomly shuffle all the paragraphs of a document.
In \emph{\mpslong{}} (\mps{}), we shuffle a subset of the paragraphs of a document. 
Precisely, we randomly pick two paragraphs from a document and shuffle all the paragraphs between them including those two as well.
For example, in the \mps{} of Figure~\ref{Fig:pc_example}, only paragraph number 3,4 and 5 are shuffled.

In \emph{\pdlong{}} (\pd{}), we drop randomly selected 30\% of the paragraphs of a document.
Figure~\ref{Fig:pc_example} shows an example of \pd{} where paragraph number 2 and 3 are dropped.

In \emph{\prsplong{}} (\prsp{}), we randomly choose two paragraphs from a document and replace all the paragraphs between them (including those two as well) with the paragraphs of another document of the same prompt. Hence, the main theme of the replaced document is still intact but the logical sequencing would be slightly distorted.
Note that, during replacement of the paragraphs, the positions of the chosen paragraphs of another document are the same as the positions of the to be replaced paragraphs of the current document. For example, if we want to replace paragraph number 3 and 4 of a document, then we choose paragraph number 3 and 4 of another document of the same prompt for replacement. 
In the \prsp{} example of Figure~\ref{Fig:pc_example}, paragraph number 3 and 4 are replaced from paragraphs of another essay of the same prompt.
Lastly, we perform a corruption called \emph{\prdplong{}} (\prdp{}) which is same as the \prsp{} but this time the paragraphs are replaced from another document of different prompt. Therefore, this corruption techniques produce incoherent documents where both main idea as well as logical sequencing are distorted. 
It is to be noted that, we hope to capture paragraph-level long range dependencies with these corruption strategies.

\begin{figure}[!t]
\centering
  \includegraphics[width=55mm, height=45mm]{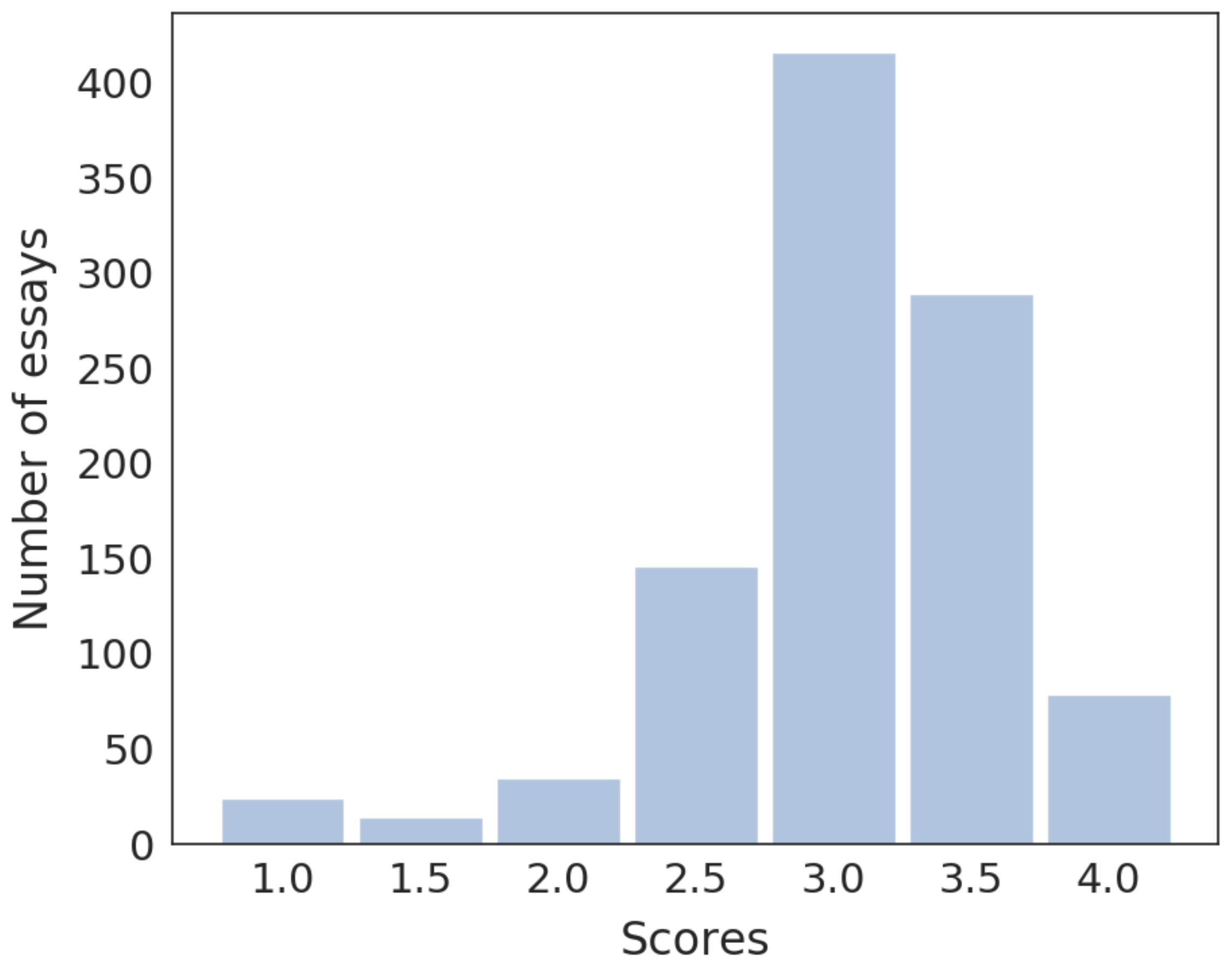}
  \captionof{figure}{Distribution of Organization scores}
  \label{Fig:score_distribution}
\end{figure}

\subsection{Discourse Corruption (DC) Pre-training}

We treat DC pre-training as a multi-class (or binary) classification task where the encoder assigns a label to each document.
In our experiments, we consider many combinations of corruption types (see Table~\ref{table:pre-training_result}).
For example, for 6-way DC pre-training, the encoder tries to predict which class the document belongs to among the 6 classes (original essays and \cps{}, \mps{}, \pd{}, \prsp{}, \prdp{} corrupted essays).
For implementation, we add a classification layer on top of the base document encoder (Section~\ref{sec:doc_enc}).
The classification layer consists of (i) a linear layer that takes $\mathbf{h}^\text{base}$ as input and (ii) a softmax layer.
To train the model parameters, we minimize the cross-entropy loss function.
% %
% \begin{align}
% \nonumber
% P_i(d) = \frac{e^{d_i}}{\sum^C_{j = 1}e^{d_j}},
% \end{align}
% \noindent
% where $d$ is the real-valued vector whose length is number of classes $C$.
%
% \begin{align}
% \nonumber \mathcal{L} = - \sum_{d \in N}\sum^C_{i = 1} \: P^g_i(d) \cdot log(P_i(d)),
% \end{align}
% \noindent
% where $N$ is the total number of documents, $C$ is the number of classes, $d$ is a document.

% Note that $y_i$ is automatically assigned in the corruption process where an original document has a label of $1$ and an artificially corrupted document has a label of $0$.

% \begin{align}
% \nonumber \mathcal{L} = - \sum^N_{i = 1} \: & y_i \text{log}(P(y(d_i) = 1 | d_i)) \: +\\
% \nonumber & (1 - y_i) \text{log}(1 - P(y(d_i) = 1 | d_i)) \:\:,
% \end{align}
% \begin{align}
% \nonumber
% P(y(d) = 1 | d) = \sigma({\bf w}^\text{unsup} \cdot {\bf h}^\text{mean}) \:\:,
% \end{align}

%\noindent
% where $y$ is a binary function mapping from a document $d$ to $\{ 0,1 \}$, in which $1$ represents the document is coherent/cohesive and $0$ represents not.
% The base document representation ${\bf h}^\text{mean}$ (Eq.~\ref{eq:base_doc}) is multiplied with a weight vector ${\bf w}^\text{unsup}$, and the sigmoid function $\sigma$ returns a probability that the given document $d$ is coherent/cohesive. 

%%%%%%%%%%%%%%%%%%%%%%%%%%%%%%%%%%%%%%%%%%%%%%%%%%%%%%%%%%%%%%%%%%%%%%%%%%%%%%%%%%%%%%%%%
%%%%%%%%%%%%%%%%%%%%%%%%%%%%%%%%%%%%%%%%%%%%%%%%%%%%%%%%%%%%%%%%%%%%%%%%%%%%%%%%%%%%%%%%%
%%%%%%%%%%%%%%%%%%%%%%%%%%%%%%%%%%%%%%%%%%%%%%%%%%%%%%%%%%%%%%%%%%%%%%%%%%%%%%%%%%%%%%%%%

\section{Experimental Setup}
\label{sec:setup}

\subsection{Data}
\subsubsection{Essay Organization Scoring}
\begin{figure}[!h]
\centering
  \includegraphics[width=65mm, height=52mm]{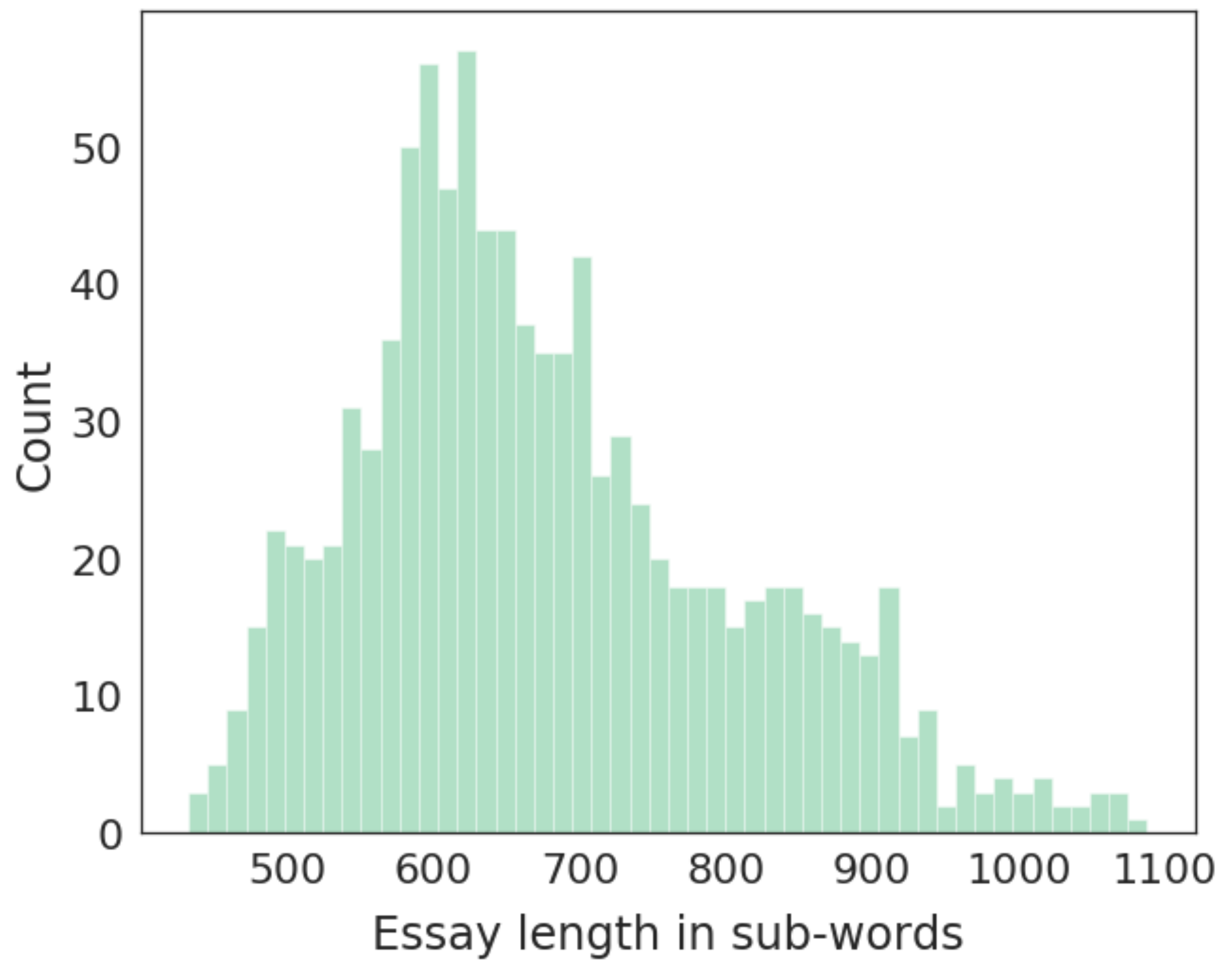}
  \captionof{figure}{Histogram of lengths of ICLE essays used in scoring}
  \label{Fig:histogram}
\end{figure}

We use the International Corpus of Learner English (ICLE) ~\cite{granger2009international} for essay scoring which contains 6,085 essays and 3.7 million words.
Most essays (91\%) are argumentative and vary in length, having 7.6 paragraphs and 33.8 sentences on average~\cite{wachsmuth2016using}. 
Some essays have been annotated with scores along multiple dimension among which 1,003 essays are annotated with Organization scores. The scores range from 1.0 (worst score) to 4.0 (best score) at half-point increments. The distribution of Organization scores is demonstrated in Figure~\ref{Fig:score_distribution}.
For our scoring task, we utilize these 1,003 essays.
The average number of tokens per esssay is 679 (in sub-words) and the longest essay has 1090 tokens. The histogram of the essay lengths is shown in Figure~\ref{Fig:histogram}.

\subsubsection{DC Pre-training}
To pre-train the document encoder, we use four datasets, (i) the Kaggle's Automated Student Assessment Prize (ASAP) dataset\footnote{\url{https://www.kaggle.com/c/asap-aes}} (12,976 essays) (ii) TOEFL11 \cite{blanchard2013toefl11} dataset (12,100 essays), (iii) The International Corpus Network of Asian Learners of English (ICNALE) \cite{ishikawa2013icnale} dataset (5,600 essays), and (iv) the ICLE essays not used for Organization scoring (4,546 essays).
Total 35,222 essays from the four datasets are used during pre-training with SC and DIC.
However, for pre-training with all types of PC, we use only 16,646 essays (TOEFL11 and ICLE essays) since ASAP and ICNALE essays have a single paragraph.

\begin{figure*}[!b]
\centering
  \includegraphics[clip, width=\textwidth]{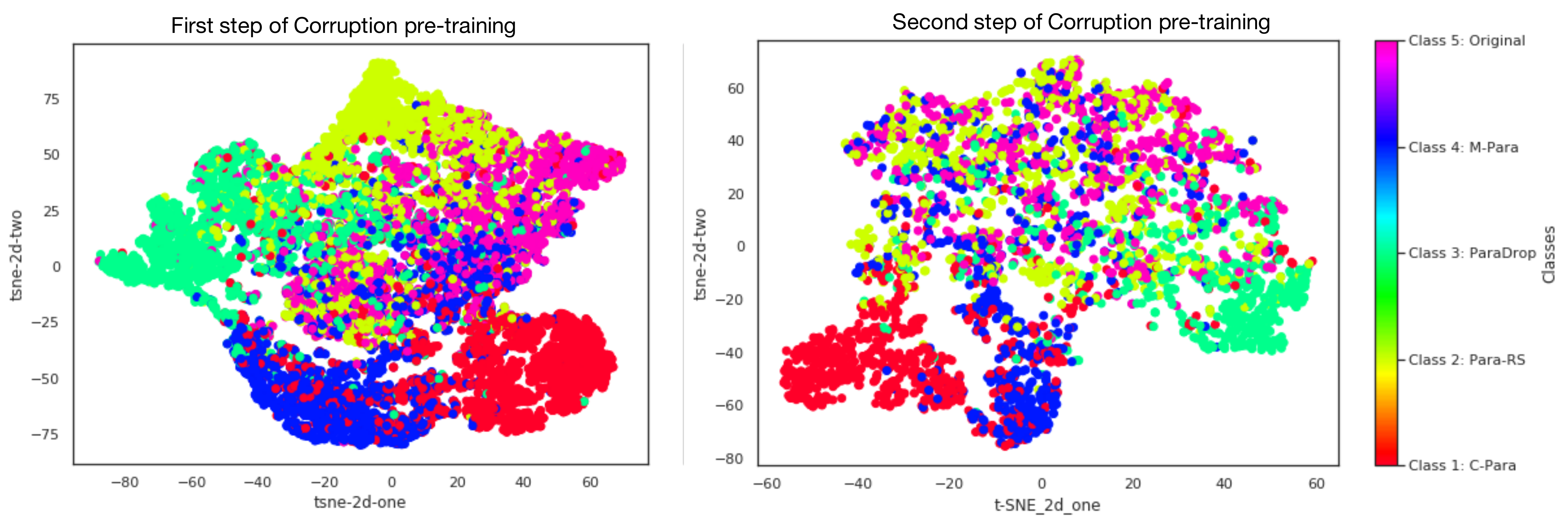}
  \caption{Visualization of document representations obtained from DC pre-trained (5-way classification scheme) encoder}
  \label{Fig:pretrain_docvec}
\end{figure*}

% \footnote{During pre-training with paragraph shuffled essays, we use only 16,646 essays (TOEFL11 and ICLE essays) since ASAP and ICNALE essays have a single paragraph.}

% See Appendix~\ref{sec:hyperp} and \ref{sec:prep} for further details on the hyperparameters and preprocessing.

\subsection{Evaluation Procedure}
We use five-fold cross-validation for evaluating our models with the same split as Persing et al. \cite{persing2010modeling} and Wachsmuth et al. \cite{wachsmuth2016using}.
However, our results are not directly comparable to them since our training data is smaller as we reserve a validation set (100 essays) for model selection while they do not.
We use the mean squared error (MSE) as an evaluation measure.
The reported results are averaged over five folds.

\begin{table*}[t!]
\centering
\footnotesize
 \begin{tabular}{lccc}
    \toprule
    Pretraining Phase & Classification Task & Corruption Type Used & Validation Accuracy \\
    \midrule
    \multirow{8}{*}{1st Step (All pre-training data)} & Binary & \css{} & 0.990 \\
    & Binary & \mss{} & 0.971 \\
    & Binary & C-DI & 0.984 \\
    & Binary & M-DI & 0.971 \\
    & Binary & C-Para & 0.919 \\
    & 3-way & C-Para, M-para & 0.786 \\
    & 4-way & C-Para, M-para, ParaDrop & 0.770 \\
    & 5-way & C-Para, M-Para, ParaDrop, Para-RS & 0.707 \\
    & 6-way & C-Para, M-Para, ParaDrop, Para-RS, Para-RD & 0.734 \\
    
    \midrule
    \multirow{8}{*}{2nd Step (Finetuned on ICLE pre-training data)} & Binary & C-Sent & 0.999 \\
    & Binary & M-Sent & 0.990 \\
    & Binary & C-DI & 1.000 \\
    & Binary & M-DI & 0.998 \\
    & Binary & C-Para & 0.890 \\
    & 3-way & C-Para, M-Para & 0.717 \\
    & 4-way & C-Para, M-Para, ParaDrop & 0.656 \\
    & 5-way & C-Para, M-Para, ParaDrop, Para-RS & 0.606 \\
    & 6-way & C-Para, M-Para, ParaDrop, Para-RS, Para-RD & 0.666 \\
    \bottomrule
    \end{tabular}
\caption{Performance of classification tasks in the first step (using large-scale unlabeled essays) and \\ second step of Corruption Pre-training (using unlabeled essays of target essay scoring corpus)}
\label{table:pre-training_result}
\end{table*}

% To investigate the performance, 
We evaluated two learning strategies of encoder in essay scoring task: \emph{fine-tuning} and \emph{fixed}.
In the fine-tuning setting, both the pre-trained base document encoder and auxiliary encoder are fine-tuned on the essay scoring task.
In the fixed setting, only the parameters of auxiliary encoder are fine-tuned.

Our baseline model is the \emph{Base+AE} model. 
In our preliminary experiments, we tried different settings such as finetune Base (pre-trained Longformer) model first then merge AE, finetune both Base and AE and then merge etc. However, we found that merging both models at the same time (either in fine-tuning or fixed encoder setting) provides the best performance.
Therefore, even for all the proposed systems, we merge the DC pre-trained Base model and AE at the same time in both fine-tuning and fixed-encoder setting.

\subsection{Preprocessing}
\label{sec:preprocessing}
%We use the same preprocessing step for both pre-training and essay scoring data.
%We lowercase the tokens %and normalize the gold-standard scores to the range of [0, 1]. 
We use the same preprocessing steps for both pre-training and essay scoring. 
We lowercase the tokens and specify an essay's paragraph boundaries with special tokens.
Special tokens $\text{[CLS]}$ and $\text{[EOS]}$ are inserted at the beginning and end of each essay respectively.
We normalize the gold-standard scores to the range of [0, 1].
During pre-training with SC and DIC, paragraph boundaries are not used.

For DIC, we collect 847 DIs from the Web.\footnote{\url{http://www.studygs.net/wrtstr6.htm}, \url{http://home.ku.edu.tr/~doregan/Writing/Cohesion.html} etc.}
We exclude the DI ``and'' since it is not always used for initiating logic (e.g., milk, banana \emph{and} tea). 
In essay scoring dataset, we found 176 DIs and around 24 DIs per essay. In the pre-training data, the total number of DIs is 204 and the average number of DIs per essay is around 13. We identified DIs by simple string-pattern matching.
% During our testing phase, we re-scale the predicted normalized scores to the original range of scores and then measure the performance.

\subsection{Implementation Choices}
\label{sec:hyperp}
From the two sizes of pre-trained Longformer models, we use Longformer-base model. The global attention of Longformer is set on the $\text{[CLS]}$ token.
For the auxiliary encoder, we use a BiLSTM with hidden units of 200 in each layer ($d^\text{AUX}=200$).

We use Adam optimizer, batch sizes of 4 on the first-step of pre-training and batch sizes of 2 on the second-step of pre-training as well as on the essay scoring.
The learning rate is set to $1e-5$ for pre-training and  fine-tuning setting of essay scoring while it is set to $0.001$ for fixed encoder setting of essay scoring.
We use early stopping with patience 12 (5 for pre-training), and train the network for 100 epochs.
In the pre-training phase, 80\% data is used for training and 20\% of the data is used for validation.
We perform hyperparameter tuning for the scoring task and choose the best model. We tuned dropout rates (0.5, 0.7, 0.9) for all models on the validation set.
To select hyper-parameters, we monitor performance on validation set and choose the model that yields the lowest MSE. 
We choose the best model for each particular fold.
In testing phase, we re-scale the predicted normalized scores to the original range of scores and then measure the performance.

\begin{table*}[ht!]
\centering
%\footnotesize
\small 
 \begin{tabular}{lccccc}
    \toprule
     Model & Classification Task & Corruption Type & Fine-tuning & \multicolumn{2}{c}{Mean Squared Error} \\
     & & & & \hspace{0.5cm} Organization \\
    \midrule
    \multirow{2}{*}{Baseline} & - & - & - & 0.175  \\
    & - & - & \checkmark & 0.181  \\
    \midrule
    \multirow{18}{*}{Proposed}  & Binary  & \css{} & & 0.188 \\ 
    & Binary  & \css{}  & \checkmark &  0.183  \\ \cline{2-5} 
    & Binary & \mss{} & & \textbf{0.174} \\ 
    & Binary & \mss{} & \checkmark & 0.179 \\ \cline{2-5}
    & Binary & \cdis{} & & 0.189 \\ 
    & Binary & \cdis{} & \checkmark & 0.185 \\  \cline{2-5} 
    & Binary & \mdis{} & & 0.183 \\
    & Binary & \mdis{} & \checkmark & 0.198 \\  \cline{2-5} 
    & Binary & \cps{} & & \textbf{0.172} \\
    & Binary & \cps{} &  \checkmark & \textbf{0.167}\textsuperscript{*} \\  \cline{2-5} 
    & 3-way & \cps{}, \mps{} & & \textbf{0.173} \\
    & 3-way & \cps{}, \mps{} &  \checkmark & \textbf{\underline{0.162}}\textsuperscript{*}  \\ \cline{2-5} 
    & 4-way & \cps{}, \mps{}, \pd{} & & \textbf{0.169}  \\
    & 4-way & \cps{}, \mps{}, \pd{} &  \checkmark & \textbf{\underline{0.157}}\textsuperscript{*}   \\ \cline{2-5}
    & 5-way  & \cps{}, \mps{}, \pd{}, \prsp{} & & \textbf{0.166}\textsuperscript{*}  \\
    & 5-way & \cps{}, \mps{}, \pd{}, \prsp{} &  \checkmark & \textbf{\underline{0.155}}\textsuperscript{*}   \\ \cline{2-5}
    & 6-way & \cps{}, \mps{}, \pd{}, \prsp{}, \prdp{} &   & 0.179  \\
    & 6-way & \cps{}, \mps{}, \pd{}, \prsp{}, \prdp{} &  \checkmark & \textbf{\underline{0.162}}\textsuperscript{*}  \\  

    \midrule
    Persing et al. (2010) & & & & 0.175  \\
    Wachsmuth et al. (2016) & & & & 0.164 \\
    \bottomrule
    \end{tabular}
\caption{Performance of essay scoring. Numbers in \textbf{bold} and \textbf{underline} denote improvement over baseline and previous state-of-the-art respectively. `*' indicates a statistical significance (Wilcoxon signed-rank test, $p < 0.05$) against the baseline models. }
\label{table:essay_scoring}
\end{table*}

%%%%%%%%%%%%%%%%%%%%%%%%%%%%%%%%%%%%%%%%%%%%%%%%%%%%%%%%%%%%%
%%%%%%%%%%%%%%%%%%%%%%%%%%%%%%%%%%%%%%%%%%%%%%%%%%%%%%%%%%%%
%%%%%%%%%%%%%%%%%%%%%%%%%%%%%%%%%%%%%%%%%%%%%%%%%%%%%%%%%%%%
%%%%%%%%%%%%%%%%%%%%%%%%%%%%%%%%%%%%%%%%%%%%%%%%%%%%%%%%%%%%

\section{Results and Discussion}
\label{Sec:results}

\subsection{Results of DC Pre-training}

Table~\ref{table:pre-training_result} shows the classification accuracy of both steps of DC pre-training on the validation data.
We see that the document encoder learns to distinguish not only between coherent/cohesive and incoherent/incohesive documents (binary classification) but also between different types of incoherent (3,4,5 and 6 way classification) documents. 
% In addtion, we also see that as we add more corruption types (i.e. increase classes), the validation accuracy decreases, which means that it becomes more difficult for the encoder to distinguish between several types of corrupted documents.

Pre-training with \cdis{} provides the best classification accuracy. 
We anticipate that since we do not change the position of the discourse indicators (DIs) during shuffling, the encoder might learn only the sequence of DIs within each essay and try to distinguish between the DI sequence of original and corrupted essays. 
Therefore, the task becomes easier for the encoder.

The visualization of document vectors obtained from the first and second step of DC pre-training (5-way classification task) is shown in Figure~\ref{Fig:pretrain_docvec}. 
To visualize the high-dimensional document vectors into a 2-dimensional space, we use dimensionality reduction algorithm T-Distributed Stochastic Neighbouring Entities (t-SNE).
Figure~\ref{Fig:pretrain_docvec} shows that the encoder is able to perfectly separate \cps{} essays from other essays since the transition of ideas between paragraphs is fully distorted in these essays, hence easy to distinguish.
We also see that the encoder well separate \mps{} and \pd{} essays compared to \prsp{} essays.
\prsp{} essays lies close to the original coherent essays and overlaps a lot.
We speculate that since we replace the paragraphs of the same positions, the sequencing of ideas of \prsp{} essays is the least distorted compared to \mps{}, \pd{} or \cps{} essays, hence these essays are similar to the original essays.

\subsection{Results of Essay Scoring}
Table~\ref{table:essay_scoring} lists MSE (averaged over five folds) of baseline model and our proposed systems (DC pre-trained) for Organization scoring task.\footnote{Our model is Base+AE model~(Section \ref{sec:doc_enc}, \ref{sec:ae}). The performance of the only Base (pre-traned Longformer) encoder without AE and without any DC pre-training when finetuned on essay Organization scoring is: MSE = 0.246}
It shows that the proposed unsupervised DC pre-training improves the performance of essay Organization scoring (statistically significant by Wilcoxon's signed rank test, $p < 0.05$) and we obtain significant performance gain over the baseline model. 
Also, we achieve new state-of-the-art result with our proposed method.

The best performance is obtained with the 5-way DC Pre-training.
These results support our hypothesis that training with corrupted documents helps a document encoder learn logical sequence-aware text representations.
In most of the cases, fine-tuning the encoder for scoring task provides better performance.

From Table~\ref{table:essay_scoring} we observe that paragraph corruption based DC pre-training is effective for Organization scoring while sentence and DI corruption based is not.
% It means that the sentence and DI level corruptions is unable to capture the global coherence.
This could be attributed to the fact that the paragraph level transition of ideas (global coherence) is not captured by sentence and DI level corruption. 
Besides, a manual inspection of DIs identified by the system shows that the identification of DIs is not always reliable.
Almost half of DIs identified by our simple pattern matching algorithm (see Section~\ref{sec:preprocessing}) were not actually DIs (e.g., \emph{we have survived \textbf{so} far only external difficulties}).
We also found that some DI-shuffled documents are often cohesive.
This happens when original document counterparts have two or more DIs with more or less same meaning (e.g., \emph{since} and \emph{because}).
% We speculate that this confuses the document encoder in the pre-training process.

It can be seen that as the classification task of Corruption Pre-training becomes more complicated by adding more corruption types, the essay scoring performance improves (except for 6-way classification).
We obtain the best performance with 5-way classification task.
We speculate that this is because with more corruption types, the model learns more styles of transition of ideas among paragraphs as well as differences between them. Finally, the model connects those differences to scores at the essay scoring phase by figuring out which flow of concepts is better than the other.

It should be noted that 6-way classification task could not outperform 5-way classification task.
This might be because of adding \prdp{} corruption in 6-way classification task. Since in \prdp{}, we replace the paragraphs of document with paragraphs of a document of different prompt, instead of learning the flow of the ideas throughout the text the encoder might also be learning something else (e.g, topic difference).
We speculate that this confuses the document encoder at the essay scoring phase.

\begin{table*}[ht!]
\centering
 %footnotesize
\small 
 \tabcolsep 0.25cm
 \begin{tabular}{lccccc}
    \toprule
     Model & Classification Task & Corruption Type & Fine-tuning & \multicolumn{2}{c}{Mean Squared Error} \\
     & & & & \hspace{0.5cm} Organization \\
    \midrule
    \multirow{2}{*}{Baseline} & - & - & - & 0.175  \\
    & - & - & \checkmark & 0.181  \\
    \midrule
    \multirow{6}{*}{Proposed}& 5-way & \cps{}, \mps{}, \pd{}, \prsp{} & & \textbf{0.166}\textsuperscript{*}  \\
    & 5-way & \cps{}, \mps{}, \pd{}, \prsp{} &  \checkmark & \textbf{0.155}\textsuperscript{*}   \\ \cline{2-5}
    & 5-way to Binary & \cps{}, \mps{}, \pd{}, \prsp{} & & 0.179  \\
    & 5-way to Binary & \cps{}, \mps{}, \pd{}, \prsp{} &  \checkmark & 0.185 \\
    & 5-way to 3-way & \cps{}, \mps{}, \pd{}, \prsp{} & & 0.181  \\
    & 5-way to 3-way & \cps{}, \mps{}, \pd{}, \prsp{} &  \checkmark & \textbf{0.162}\textsuperscript{*} \\
    \bottomrule
    \end{tabular}
\caption{Essay scoring results when a 5-way DC pre-trained model is transformed to a Binary and 3-way DC pre-trained model}
\label{table:tranform_class}
\end{table*}

\subsection{Analysis}

\subsubsection{Importance of Fine-grained Corruption Types}

To investigate how important for the model to learn the difference between fine-grained corruption types, we collapsed four corruption types into one or two classes in DC pre-training.
Specifically, we reduced the best performing 5-way DC pre-training to (i) binary DC pre-training with original v.s. corrupted essays (\{\cps{}, \mps{}, \pd{}, \prsp{}\}), and to (ii) 3-way DC pre-training  with original v.s. fully corrupted (\cps) v.s. partially corrupted essays (\{\mps{}, \pd{}, \prsp{}\}).

Table~\ref{table:tranform_class} demonstrates the results.
It shows that transforming a 5-way to a binary classification provides even worse results than the baseline.
We think this is because we combined fully corrupted (CPS) essays with partially corrupted (MPS, PD, PRSP) essays, so the model cannot distinguish between very bad and relatively bad essays.
This hypothesis is proved when we transform it to a 3-way classification task.
We obtain much better performance during finetuning but not as good as the original 5-way classification task.
Overall, these experiments indicate that differentiating between fine-grained corruption types is essential.

\begin{figure}[!b]
\centering
  \includegraphics[height=60mm]{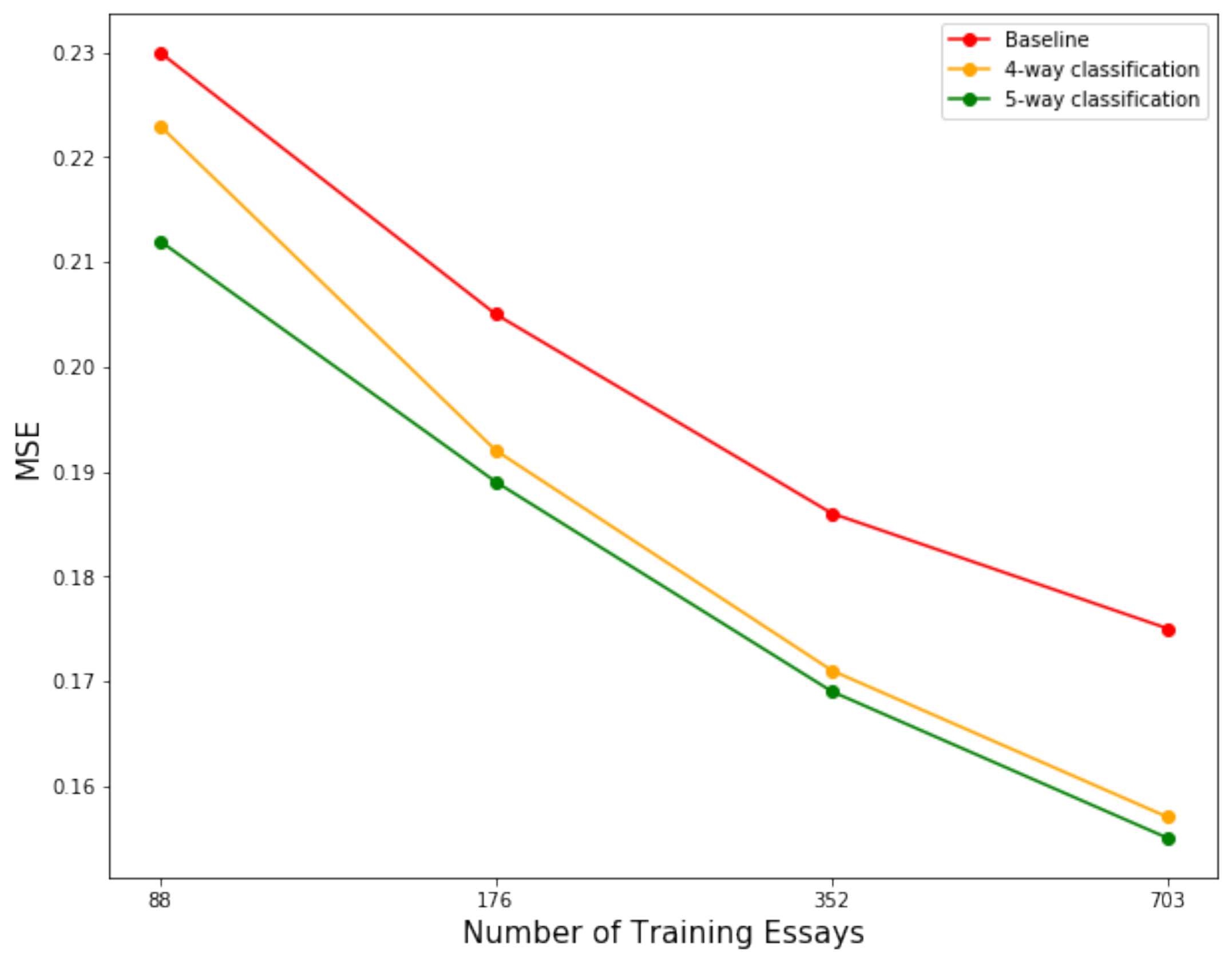}
  \caption{Plot of training data size vs MSE at essay scoring phase}
  \label{Fig:data_vs_mse}
\end{figure}

\subsubsection{Effectiveness of Corruption Pre-training in Low Resource Setting}

To investigate how beneficial our DC pre-training is when labeled data is less available, we reduce the training data at the essay scoring phase.
We examine two best performing DC pre-trained models (4-way and 5-way classification) and compare them with the baseline model (model without DC pre-training).

Figure~\ref{Fig:data_vs_mse} shows a plot of number of training essays vs. MSE.
MSE is obtained with all training data (703 essays) as well as with training data being reduced to $\frac{1}{2}$ (352 essays), $\frac{1}{4}$ (176 essays) and $\frac{1}{8}$ (88 essays). 
We see that our proposed models constantly outperform the baseline model when we reduce the training data.
It indicates the strength and effectiveness of our DC pre-training even with less information from labeled data and that the model understands which Organization structure is better than the others.

% Our 4-way classification model (orange line) does not perform very well when we reduce training data to $\frac{1}{4}$ and $\frac{1}{8}$.
% However, our best model (5-way classification) continuously performs the baseline model with similar performance gap even when we reduce the training data to $\frac{1}{8}$. This result specifies that having more fine-grained corruption types in Corruption Pre-training helps the model to be less dependent on the annotated information of which essay Organization is better.

Our 4-way DC model (orange line) does not perform better than 5-way DC model (green line).
This result indicates that having more fine-grained corruption types in DC pre-training helps the model to be less dependent on the annotated information of which essay Organization is better.

\subsubsection{Essay Embeddings}

In order to identify which scores are better distinguished by our models than by the baseline model, we visualized essay embeddings (i.e. $\mathbf{h}^\text{base}$) obtained from the fine-tuned baseline model and our proposed DC pre-trained (5-way classification) model.

\begin{figure*}
\centering
  \includegraphics[height=60mm]{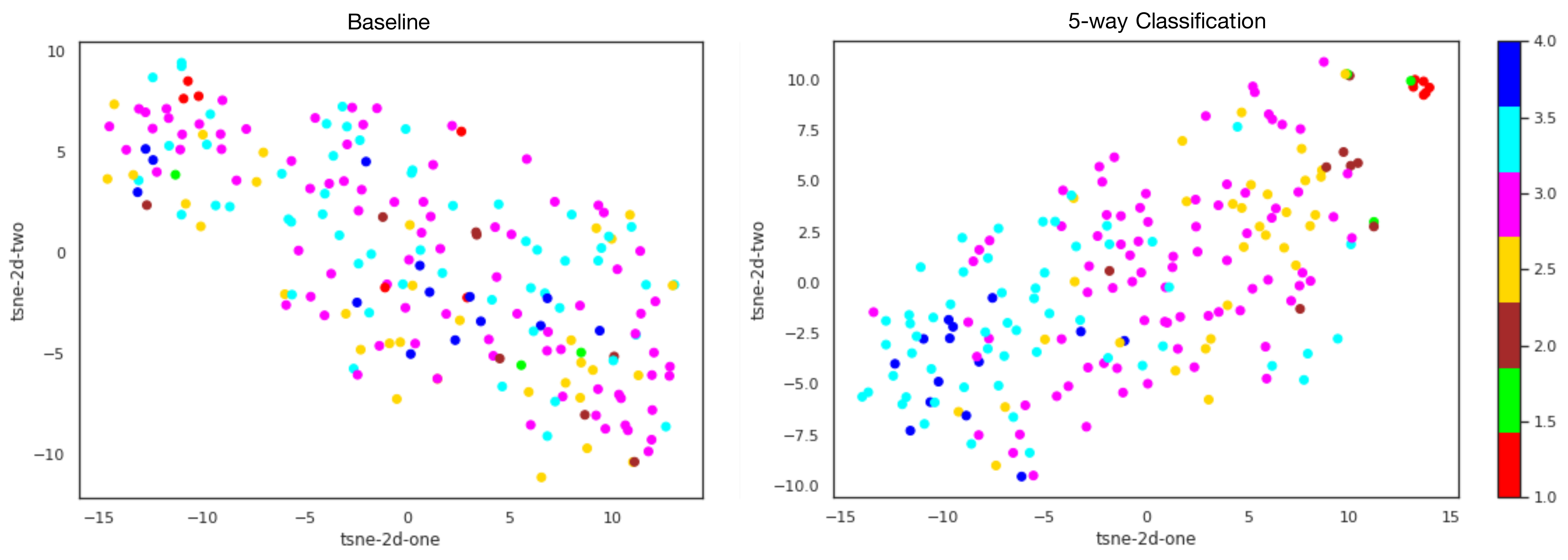}
  \caption{Visualization of essay representations}
  \label{Fig:docvec_viz}
\end{figure*}

The results are shown in Figure~\ref{Fig:docvec_viz}.
In the baseline model essay embeddings, the essays are scattered, and the low-scored essays (scored 1, red dots) are sometimes close to the high-scored essays (scored 4, blue dots) (upper-left of the figure).
In contrast, the essay representations of our DC pre-training (5-way classification) shows that our model is good at separating essays of different scores and that more cluster of scores appear compared to the baseline model. 
The highest scored (scored 4, blue dots) and the lowest scored (scored 1, red dots) essays are at the complete opposite position and furthest from each other in the embedding space.
This means our model knows the difference between high schored and low scored Organization.
We see that the lowest scored essays (red dots) are clustered and fully separated from other essays. 
Besides, other low scored essays (scored 1.5 and 2.0, lime and brown dots respectively) as well as highest scored essays (scored 4, blue dots) are also well distinguished.
This represents that our model is not only good at separating very bad Organization from very good ones but also good at distinguishing different levels of ``goodness" of essay Organization.
% Besides, other low scored essays (scored 1.5 and 2.0, lime and brown dots respectively) are close to each other and far from high scored essays. The high scored essays (blue and cyan dots) are close to each other and the medium scored essays (gold and magenta dots) resides in the middle of high and low scored essays in the embedding space.

Table~\ref{table:improved_scores} presents 10 test instances for which the prediction of our DC pre-trained model is better (i.e., lower MSE between gold and predicted score) than the baseline model.
Column 1 shows the gold essay score, column 2 and 3 shows the scores predicted by the baseline model and our best DC pre-trained model (5-way classification) respectively\footnote{The predicted scores are shown till one decimal point}.
Column 4 presents the MSE between gold score and baseline predicted score whereas column 5 presents the MSE between gold score and score predicted by DC pre-trained model. 
% Generally, MSE indicates the gap between gold and predicted score, the lower the MSE, the closer the gold and predicted scores are.
Table~\ref{table:improved_scores} shows that our DC pre-trained model is specially good at predicting low-to-medium and high essay scores when compared to the baseline. If we look at the MSE difference between column 4 and 5, we see how better DC pre-trained model's prediction is than the baseline. %Hence, it can be said that our DC Pre-training is substantially effective in distinguishing bad Organization structure.

% The results indicate that paragraph shuffling is the most effective in both scoring tasks (statistically significant by Wilcoxon's signed rank test, $p < 0.05$).
%This could be attributed to the fact that both scores are closely related to paragraph structure.
% This could be attributed to the fact that paragraph sequences create a more clear organizational and argumentative structure.
% Suppose that an essay first introduces a topic, states their position, supports their position and then concludes.
% Then, the structure of the essay would be regarded as ``well-organized''. 
% Moreover,  the argument of the essay would be considered ``strong'' since it provides support for their position.
% The results suggest that such levels of abstractions (e.g., \emph{Introduction-Body-Body-Conclusion}) are well captured at a paragraph-level, but not at a sentence-level or DI-level alone.

%%%%%%%%%%%%%%%%%%%%%%%%%%%%%%%%%%%%%%%%%%%%%%%%%%%%%%%%%%%%%%%%%%%%%%%%%%%%%%%%%%%%%%%%%
%%%%%%%%%%%%%%%%%%%%%%%%%%%%%%%%%%%%%%%%%%%%%%%%%%%%%%%%%%%%%%%%%%%%%%%%%%%%%%%%%%%%%%%%%
%%%%%%%%%%%%%%%%%%%%%%%%%%%%%%%%%%%%%%%%%%%%%%%%%%%%%%%%%%%%%%%%%%%%%%%%%%%%%%%%%%%%%%%%%

\section{Conclusion}
% NLP tasks have been greatly benefited from different unsupervised learning approaches. 
% Although discourse structure is an important facet of text documents, less studies have been conducted to capture discourse structure in an unsupervised manner. 
% Most of the document representation learning depend on parsers or discourse annotated data for capturing discourse structure. 
% However, parsers (e.g, discourse parser, argumentation parser etc.) are computationally expensive and the performance of the parsers is not always adequate. 
% Furthermore, obtaining large-scale labeled dataset is costly and creating labeled dataset requires strict annotation procedure which vary from dataset to dataset. 
% Hence, creating a large labeled corpora for some particular task is sometimes infeasible. 
% In this paper, we address this issue by utilizing unlabeled data for the incorporation of discourse structure in document representation.

In this paper, we proposed an unsupervised pre-training strategy to capture discourse structure (i.e., coherence and cohesion) of essay Organization.
We have presented several token, sentence and paragraph level corruption techniques that produce several types of fully corrupted (totally incoherent/incohesive) or partially corrupted (partially incoherent/incohesive) essays.
Then we train a document encoder to discriminate between original and their artificially corrupted essays in order to make the encoder logical-sequence aware. 
After that, the logical-sequence aware encoder is used to obtain feature vectors of essays for the task of essay Organization scoring.
Our proposed pre-training strategy does not require any parser or annotation.
The experimental results show that the proposed method successfully captures the discourse structures of essay Organization and we obtain new state-of-the art result for essay Organization scoring. 
It also shows that the combination of MLM pre-trained document encoder and paragraph level discourse corruption pre-training is effective to capture the discourse of essay Organization. 
The combination of these two can track both global and local coherence.

\begin{table}[!t]
\centering
\footnotesize
 \begin{tabular}{lcccc}
    \toprule
    \shortstack{Gold \\ Score} & \shortstack{Baseline \\ Predicted} & \shortstack{5-way \\ Predicted} & \shortstack{MSE \\ (gold\&baseP)} & \shortstack{MSE \\ (gold\&5-wayP)} \\
    %  \midrule
    % Gold Score & Baseline predicted & 5-way classification predicted & MSE (gold\&baseline) & MSE (gold\&5-way)\\
    \midrule
    1.0 & 2.3 & 1.1 & 1.69 & 0.01 \\
    2.5 & 1.2 & 2.1 & 1.69 & 0.16 \\
    2.5 & 3.7 & 2.5 & 1.44 & 0.00 \\
    2.5 & 1.4 & 2.5 & 1.21 & 0.00 \\
    1.0 & 2.3 & 1.7 & 1.69 & 0.49 \\
    2.0 & 3.3 & 2.9 & 1.69 & 0.81 \\
    %3.0 & 2.1 & 3.0 & 0.81 & 0.00 \\
    %2.5 & 3.4 & 2.6 & 0.81 & 0.01 \\
    %2.5 & 3.5 & 3.0 & 1.00 & 0.25 \\
    2.0 & 2.9 & 2.4 & 0.81 & 0.09 \\
    4.0 & 3.1 & 3.6 & 0.81 & 0.16 \\
    4.0 & 3.2 & 3.7 & 0.64 & 0.09 \\
    1.5 & 2.5 & 2.2 & 1.00 & 0.49 \\
    
    \bottomrule
    \end{tabular}
\caption{Score prediction of test instances by baseline \\ and our best DC  pre-trained model}
\label{table:improved_scores}
\end{table}

%Our analysis show that DC pre-trained model is specially good at distinguishing bad Organization structure since we create examples of bad Organization by different types of corruption.

One possible future direction of this work could be figuring out how to exploit other unannotated argumentative texts (except student essays) for the proposed pre-training method. Since student essays are not perfect (have grammatical or spelling error), it would be interesting to see how the proposed method behaves when pre-trained with perfectly written or error-less texts.
% in this scenario.
We hope that this work would inspire the exploration of new ways of unsupervised encapsulation of discourse structure in text representation.

\ifCLASSOPTIONcompsoc
  \section*{Acknowledgments}
\else
  \section*{Acknowledgment}
\fi

This work was supported by JSPS KAKENHI
Grant Number 19K20332.

\ifCLASSOPTIONcaptionsoff
  \newpage
\fi

\bibliographystyle{IEEEtran}
\bibliography{ieee.bib}

\begin{thebibliography}{10}
\providecommand{\url}[1]{#1}
\csname url@samestyle\endcsname
\providecommand{\newblock}{\relax}
\providecommand{\bibinfo}[2]{#2}
\providecommand{\BIBentrySTDinterwordspacing}{\spaceskip=0pt\relax}
\providecommand{\BIBentryALTinterwordstretchfactor}{4}
\providecommand{\BIBentryALTinterwordspacing}{\spaceskip=\fontdimen2\font plus
\BIBentryALTinterwordstretchfactor\fontdimen3\font minus
  \fontdimen4\font\relax}
\providecommand{\BIBforeignlanguage}[2]{{%
\expandafter\ifx\csname l@#1\endcsname\relax
\typeout{** WARNING: IEEEtran.bst: No hyphenation pattern has been}%
\typeout{** loaded for the language `#1'. Using the pattern for}%
\typeout{** the default language instead.}%
\else
\language=\csname l@#1\endcsname
\fi
#2}}
\providecommand{\BIBdecl}{\relax}
\BIBdecl

\bibitem{attali2006automated}
Y.~Attali and J.~Burstein, ``Automated essay scoring with
  e-rater{\textregistered} v. 2,'' \emph{The Journal of Technology, Learning
  and Assessment}, vol.~4, no.~3, 2006.

\bibitem{persing2010modeling}
I.~Persing, A.~Davis, and V.~Ng, ``{Modeling organization in student essays},''
  in \emph{Proceedings of the 2010 Conference on EMNLP}, 2010, pp. 229--239.

\bibitem{persing2013modeling}
I.~Persing and V.~Ng, ``Modeling thesis clarity in student essays,'' in
  \emph{Proceedings of the 51st Annual Meeting of the Association for
  Computational Linguistics}, 2013, pp. 260--269.

\bibitem{persing2014modeling}
I.~Persing and V.~Ng, ``Modeling prompt adherence in student essays,'' in
  \emph{Proceedings of the 52nd Annual Meeting of the Association for
  Computational Linguistics}, 2014, pp. 1534--1543.

\bibitem{persing2015modeling}
I.~Persing and V.~Ng, ``{Modeling argument strength in student essays},'' in
  \emph{Proceedings of the 53rd Annual Meeting of the ACL the 7th International
  Joint Conference on Natural Language Processing}, 2015, pp. 543--552.

\bibitem{persing2016modeling}
I.~Persing and V.~Ng, ``Modeling stance in student essays,'' in
  \emph{Proceedings of the 54th Annual Meeting of the Association for
  Computational Linguistics (Volume 1: Long Papers)}, 2016, pp. 2174--2184.

\bibitem{wachsmuth2016using}
H.~Wachsmuth, K.~Al~Khatib, and B.~Stein, ``{Using argument mining to assess
  the argumentation quality of essays},'' in \emph{Proceedings of COLING 2016,
  the 26th International Conference on Computational Linguistics: Technical
  Papers}, 2016, pp. 1680--1691.

\bibitem{mathias2018thank}
S.~Mathias and P.~Bhattacharyya, ``Thank “goodness”! a way to measure style
  in student essays,'' in \emph{Proceedings of the 5th Workshop on Natural
  Language Processing Techniques for Educational Applications}, 2018.

\bibitem{mim2019unsupervised}
F.~S. Mim, N.~Inoue, P.~Reisert, H.~Ouchi, and K.~Inui, ``Unsupervised learning
  of discourse-aware text representation for essay scoring,'' in
  \emph{Proceedings of the 57th Annual Meeting of the Association for
  Computational Linguistics: Student Research Workshop}, 2019, pp. 378--385.

\bibitem{halliday1994introduction}
M.~A. Halliday, ``{An introduction to functional grammar 2nd edition},''
  \emph{London: Arnold}, 1994.

\bibitem{barzilay2008modeling}
R.~Barzilay and M.~Lapata, ``Modeling local coherence: An entity-based
  approach,'' \emph{Computational Linguistics}, vol.~34, no.~1, pp. 1--34,
  2008.

\bibitem{unger2006genre}
C.~Unger, \emph{Genre, relevance and global coherence: The pragmatics of
  discourse type}.\hskip 1em plus 0.5em minus 0.4em\relax Springer, 2006.

\bibitem{zhang2011sentence}
R.~Zhang, ``Sentence ordering driven by local and global coherence for summary
  generation,'' in \emph{Proceedings of the ACL 2011 Student Session}, 2011,
  pp. 6--11.

\bibitem{stab2014annotating}
C.~Stab and I.~Gurevych, ``Annotating argument components and relations in
  persuasive essays,'' in \emph{Proceedings of COLING 2014, the 25th
  international conference on computational linguistics: Technical papers},
  2014, pp. 1501--1510.

\bibitem{stab2014identifying}
C.~Stab and I.~Gurevych, ``Identifying argumentative discourse structures in
  persuasive essays,'' in \emph{Proceedings of the 2014 Conference on Empirical
  Methods in Natural Language Processing (EMNLP)}, 2014.

\bibitem{mann1988rhetorical}
W.~C. Mann and S.~A. Thompson, ``Rhetorical structure theory: Toward a
  functional theory of text organization,'' \emph{Text}, vol.~8, no.~3, pp.
  243--281, 1988.

\bibitem{ji2017neural}
Y.~Ji and N.~Smith, ``{Neural discourse structure for text categorization},''
  \emph{arXiv preprint arXiv:1702.01829}, 2017.

\bibitem{adhikari2019docbert}
A.~Adhikari, A.~Ram, R.~Tang, and J.~Lin, ``Docbert: Bert for document
  classification,'' \emph{arXiv preprint arXiv:1904.08398}, 2019.

\bibitem{zhang2019hibert}
X.~Zhang, F.~Wei, and M.~Zhou, ``Hibert: Document level pre-training of
  hierarchical bidirectional transformers for document summarization,''
  \emph{arXiv preprint arXiv:1905.06566}, 2019.

\bibitem{xu2020discourse}
J.~Xu, Z.~Gan, Y.~Cheng, and J.~Liu, ``Discourse-aware neural extractive text
  summarization,'' in \emph{Proceedings of the 58th Annual Meeting of the
  Association for Computational Linguistics}, 2020, pp. 5021--5031.

\bibitem{steimeltowards}
K.~Steimel and B.~Riordan, ``Towards instance-based content scoring with
  pre-trained transformer models.''

\bibitem{liu2019automated}
J.~Liu, Y.~Xu, and Y.~Zhu, ``Automated essay scoring based on two-stage
  learning,'' \emph{arXiv preprint arXiv:1901.07744}, 2019.

\bibitem{nadeem2019automated}
F.~Nadeem, H.~Nguyen, Y.~Liu, and M.~Ostendorf, ``Automated essay scoring with
  discourse-aware neural models,'' in \emph{Proceedings of the Fourteenth
  Workshop on Innovative Use of NLP for Building Educational Applications},
  2019, pp. 484--493.

\bibitem{vaswani2017attention}
A.~Vaswani, N.~Shazeer, N.~Parmar, J.~Uszkoreit, L.~Jones, A.~N. Gomez,
  {\L}.~Kaiser, and I.~Polosukhin, ``Attention is all you need,'' in
  \emph{Advances in neural information processing systems}, 2017, pp.
  5998--6008.

\bibitem{devlin2018bert}
J.~Devlin, M.-W. Chang, K.~Lee, and K.~Toutanova, ``Bert: Pre-training of deep
  bidirectional transformers for language understanding,'' \emph{arXiv preprint
  arXiv:1810.04805}, 2018.

\bibitem{liu2019roberta}
Y.~Liu, M.~Ott, N.~Goyal, J.~Du, M.~Joshi, D.~Chen, O.~Levy, M.~Lewis,
  L.~Zettlemoyer, and V.~Stoyanov, ``Roberta: A robustly optimized bert
  pretraining approach,'' \emph{arXiv preprint arXiv:1907.11692}, 2019.

\bibitem{Beltagy2020Longformer}
I.~Beltagy, M.~E. Peters, and A.~Cohan, ``Longformer: The long-document
  transformer,'' \emph{arXiv:2004.05150}, 2020.

\bibitem{larkey1998automatic}
L.~S. Larkey, ``Automatic essay grading using text categorization techniques,''
  in \emph{Proceedings of the 21st annual international ACM SIGIR conference on
  Research and development in information retrieval}, 1998.

\bibitem{chen2013automated}
H.~Chen and B.~He, ``Automated essay scoring by maximizing human-machine
  agreement,'' in \emph{Proceedings of the 2013 Conference on Empirical Methods
  in Natural Language Processing}, 2013, pp. 1741--1752.

\bibitem{phandi2015flexible}
P.~Phandi, K.~M.~A. Chai, and H.~T. Ng, ``Flexible domain adaptation for
  automated essay scoring using correlated linear regression,'' in
  \emph{Proceedings of the 2015 Conference on Empirical Methods in Natural
  Language Processing}, 2015, pp. 431--439.

\bibitem{taghipour2016neural}
K.~Taghipour and H.~T. Ng, ``{A neural approach to automated essay scoring},''
  in \emph{Proceedings of the 2016 Conference on EMNLP}, 2016, pp. 1882--1891.

\bibitem{alikaniotis2016automatic}
D.~Alikaniotis, H.~Yannakoudakis, and M.~Rei, ``Automatic text scoring using
  neural networks,'' \emph{arXiv preprint arXiv:1606.04289}, 2016.

\bibitem{dong2016automatic}
F.~Dong and Y.~Zhang, ``Automatic features for essay scoring--an empirical
  study,'' in \emph{Proceedings of the 2016 Conference on Empirical Methods in
  Natural Language Processing}, 2016, pp. 1072--1077.

\bibitem{dong2017attention}
F.~Dong, Y.~Zhang, and J.~Yang, ``Attention-based recurrent convolutional
  neural network for automatic essay scoring,'' in \emph{Proceedings of the
  21st Conference on Computational Natural Language Learning (CoNLL 2017)},
  2017, pp. 153--162.

\bibitem{riordan2017investigating}
B.~Riordan, A.~Horbach, A.~Cahill, T.~Zesch, and C.~Lee, ``Investigating neural
  architectures for short answer scoring,'' in \emph{Proceedings of the 12th
  Workshop on Innovative Use of NLP for Building Educational Applications},
  2017, pp. 159--168.

\bibitem{farag2018neural}
Y.~Farag, H.~Yannakoudakis, and T.~Briscoe, ``{Neural automated essay scoring
  and coherence modeling for adversarially crafted input},'' \emph{arXiv
  preprint arXiv:1804.06898}, 2018.

\bibitem{zhang2018co}
H.~Zhang and D.~Litman, ``Co-attention based neural network for
  source-dependent essay scoring,'' in \emph{Proceedings of the Thirteenth
  Workshop on Innovative Use of NLP for Building Educational Applications},
  2018.

\bibitem{wang2018automatic}
Y.~Wang, Z.~Wei, Y.~Zhou, and X.-J. Huang, ``Automatic essay scoring
  incorporating rating schema via reinforcement learning,'' in
  \emph{Proceedings of the 2018 Conference on Empirical Methods in Natural
  Language Processing}, 2018, pp. 791--797.

\bibitem{cummins2018neural}
R.~Cummins and M.~Rei, ``Neural multi-task learning in automated assessment,''
  \emph{arXiv preprint arXiv:1801.06830}, 2018.

\bibitem{higgins2004evaluating}
D.~Higgins, J.~Burstein, D.~Marcu, and C.~Gentile, ``Evaluating multiple
  aspects of coherence in student essays,'' in \emph{Proceedings of the Human
  Language Technology Conference of the North American Chapter of the
  Association for Computational Linguistics: HLT-NAACL 2004}, 2004, pp.
  185--192.

\bibitem{mesgar2018neural}
M.~Mesgar and M.~Strube, ``{A Neural Local Coherence Model for Text Quality
  Assessment},'' in \emph{Proceedings of the 2018 Conference on EMNLP}, 2018,
  pp. 4328--4339.

\bibitem{mim2019unsupervisedanlp}
F.~S. Mim, N.~Inoue, P.~Reisert, H.~Ouchi, and K.~Inui, ``Unsupervised learning
  of discourse-aware text representation,'' 2019.

\bibitem{le2014distributed}
Q.~Le and T.~Mikolov, ``{Distributed representations of sentences and
  documents},'' in \emph{International Conference on Machine Learning}, 2014,
  pp. 1188--1196.

\bibitem{wu2018word}
L.~Wu, I.~E. Yen, K.~Xu, F.~Xu, A.~Balakrishnan, P.-Y. Chen, P.~Ravikumar, and
  M.~J. Witbrock, ``{Word Mover's Embedding: From Word2Vec to Document
  Embedding},'' \emph{arXiv preprint arXiv:1811.01713}, 2018.

\bibitem{Ionescu2019VectorOL}
R.~T. Ionescu and A.~M. Butnaru, ``Vector of locally-aggregated word embeddings
  (vlawe): A novel document-level representation,'' in \emph{NAACL-HLT}, 2019.

\bibitem{gupta2020psif}
V.~Gupta, A.~Saw, P.~Nokhiz, P.~Netrapalli, P.~Rai, and P.~Talukdar, ``P-sif:
  Document embeddings using partition averaging,'' in \emph{Proceedings of the
  AAAI Conference on Artificial Intelligence}, 2020.

\bibitem{chang2019language}
M.-W. Chang, K.~Toutanova, K.~Lee, and J.~Devlin, ``Language model pre-training
  for hierarchical document representations,'' \emph{arXiv preprint
  arXiv:1901.09128}, 2019.

\bibitem{peters2018deep}
M.~E. Peters, M.~Neumann, M.~Iyyer, M.~Gardner, C.~Clark, K.~Lee, and
  L.~Zettlemoyer, ``Deep contextualized word representations,'' \emph{arXiv
  preprint arXiv:1802.05365}, 2018.

\bibitem{lstm}
M.~Schuster and K.~K. Paliwal, ``{Bidirectional recurrent neural networks},''
  \emph{IEEE Transactions on Signal Processing}, vol.~45, no.~11, pp.
  2673--2681, 1997.

\bibitem{granger2009international}
S.~Granger, E.~Dagneaux, F.~Meunier, and M.~Paquot, ``{International corpus of
  learner English},'' 2009.

\bibitem{blanchard2013toefl11}
D.~Blanchard, J.~Tetreault, D.~Higgins, A.~Cahill, and M.~Chodorow, ``{TOEFL11:
  A corpus of non-native English},'' \emph{ETS Research Report Series}, vol.
  2013, no.~2, pp. i--15, 2013.

\bibitem{ishikawa2013icnale}
S.~Ishikawa, ``{ICNALE: the international corpus network of Asian learners of
  English},'' \emph{Retrieved on November}, vol.~21, p. 2014, 2013.

\end{thebibliography}

%%%%%%%%%%%%%%%%%%%%%%%%%%%%%%%%%%%%%%%%%%%%%%%%%%%%%%%%%%%%%%%%%%%%%%%%%%%%%%%%%%%%%%%%%
%%%%%%%%%%%%%%%%%%%%%%%%%%%%%%%%%%%%%%%%%%%%%%%%%%%%%%%%%%%%%%%%%%%%%%%%%%%%%%%%%%%%%%%%%
%%%%%%%%%%%%%%%%%%%%%%%%%%%%%%%%%%%%%%%%%%%%%%%%%%%%%%%%%%%%%%%%%%%%%%%%%%%%%%%%%%%%%%%%%

%\appendix

%\vspace*{-2\baselineskip}
\vspace*{-1.4cm}

\begin{IEEEbiography}
    [{\includegraphics[width=1in,height=1.25in,keepaspectratio]{biography/mim.pdf}}]{Farjana Sultana Mim} received her BSc degree in Computer Science and Engineering from Patuakhali Science and Technology University, Patuakhali, Bangladesh in 2016 and MS degree from Tohoku University, Sendai, Japan in 2019. She is currently a PhD student in Tohoku University, Sendai, Japan. Her research interest includes essay scoring, unsupervised learning, discourse analysis, argumentation and commonsense reasoning.
\end{IEEEbiography}

\vspace*{-1.2cm}

\begin{IEEEbiography}[{\includegraphics[width=1in,height=1.25in,keepaspectratio]{biography/naoya.pdf}}]{Naoya Inoue}
received the M.S. degree in engineering from the Nara Institute of Science and Technology in 2010 and the Ph.D. degree in information science
from Tohoku University in 2013.
He joined DENSO Corporation as a researcher in 2013 and had been an assistant professor at Tohoku University since 2015.
He is currently a postdoctoral associate at Stony Brook University since 2020.
His research interests include explainable QA systems and neuro-symbolic reasoning.
\end{IEEEbiography}

\vspace*{-1.2cm}

\begin{IEEEbiography}[{\includegraphics[width=1in,height=1.25in,clip,keepaspectratio]{biography/paul.pdf}}]{Paul Reisert} received his B.S. in Computer Science from Purdue University in West Lafayette, Indiana in 2010 and his M.S. and Ph.D. degrees in System Information Sciences from Tohoku University in Sendai, Japan in 2017. He is currently a post-doctoral researcher at RIKEN, Japan. His research interests include argumentation mining, discourse analysis, and natural language processing.
\end{IEEEbiography}

\vspace*{-1.2cm}

\begin{IEEEbiography}
[{\includegraphics[width=1in,height=1.25in,clip,keepaspectratio]{biography/ouchi.pdf}}]{Hiroki Ouchi} received his B.A. degree from Miyagi University of Education, Japan, in 2011, his M.A.degree from Ritsumeikan  University, Japan, in 2013, and his M.S. and Ph.D. degrees in Engineering from the Nara Institute of Science and Technology, Japan in 2015 and 2018. He is currently a post-doctoral researcher at RIKEN, Japan. His research interests include intersection of natural language processing and machine learning, in particular, structured prediction and instance-based learning.
\end{IEEEbiography}

\vspace*{-1.2cm}

\begin{IEEEbiography}[{\includegraphics[width=1in,height=1.25in,clip,keepaspectratio]{biography/inui.pdf}}]{Kentaro Inui} is a Professor of the Graduate School of Information Sciences at Tohoku University, where he is head of the Natural Language Processing Lab. He also leads the Natural Language Understanding Team at the RIKEN Center for the Advanced Intelligence Project. His research areas include natural language processing and artificial intelligence. He currently serves as Vice-chairperson of Association for Natural Language Processing, Member of Science Council of Japan, and Director of NPO FactCheck Initiative Japan.
\end{IEEEbiography}

\end{document}